\documentclass[10pt,twocolumn,letterpaper]{article}

\usepackage[pagenumbers]{wacv}               
\usepackage{multirow}
\usepackage{comment}
\usepackage{xcolor}
\usepackage{marvosym}

\definecolor{wacvblue}{rgb}{0.21,0.49,0.74}
\usepackage[pagebackref,breaklinks,colorlinks,allcolors=wacvblue]{hyperref}

\definecolor{deltapos}{HTML}{1F6FB2} 
\definecolor{deltaneg}{HTML}{C26A1B} 
\newcommand{\dpos}[1]{\textcolor{deltapos}{#1}}
\newcommand{\dneg}[1]{\textcolor{deltaneg}{#1}}

\usepackage{placeins}

\usepackage{graphicx}

\newcommand{\bnimg}[1]{\raisebox{-0.35ex}{\includegraphics[height=1.5ex]{res/img/bn_#1}}}

\PassOptionsToPackage{table}{xcolor}
\usepackage[most]{tcolorbox}   
\usepackage{tabularx}
\definecolor{titlebar}{HTML}{6B7A3A}    
\lstdefinestyle{prompt}{
    basicstyle=\ttfamily\scriptsize, breaklines=true, breakindent=0pt,
    columns=fullflexible, moredelim=**[is][\color{black}\bfseries]{|}{|},
    escapeinside={(*}{*)},
}
\newtcblisting{promptbox}[1]{
    listing only, enhanced, arc=2pt, boxrule=0.6pt,
    colback=white, colframe=titlebar, coltitle=white, colbacktitle=titlebar,
    fonttitle=\bfseries\small, title={#1},
    left=4pt, right=4pt, top=3pt, bottom=3pt,
    listing options={style=prompt},
}
\newtcolorbox{cardbox}[1]{
    enhanced, arc=2pt, boxrule=0.6pt,
    colback=white, colframe=titlebar, coltitle=white, colbacktitle=titlebar,
    fonttitle=\bfseries\small, title={#1},
    left=5pt, right=5pt, top=3pt, bottom=3pt,
}

\usepackage{dsfont}
\usepackage{amssymb}
\usepackage{wasysym}

\def\wacvPaperID{1288}
\def\confName{WACV}
\def\confYear{2027}
\title{Khondo: A Multimodal Benchmark for \\ Document Packet Splitting of Bangla Forms}
\author{%
\begin{minipage}{0.98\textwidth}
  \centering
  Abu~Tyeb~Azad \textsuperscript{1,\,(\Letter)},
  Fahim~Ahmed \textsuperscript{2},
  Ishita~Sur~Apan \textsuperscript{2},
  Ezharuddin~Jubaer \textsuperscript{2},\\
  Sumaiya~Karim~Katha \textsuperscript{2},
  Armun~Alam \textsuperscript{2},
  Amin~Ahsan~Ali \textsuperscript{2},
  Aman~Chadha \textsuperscript{3,\textdagger,\textdaggerdbl},\\
  Md~Mofijul~Islam \textsuperscript{3,\textdagger,\textdaggerdbl},
  AKM~Mahbubur~Rahman \textsuperscript{2,\textdagger}\\[6pt]
  \resizebox{\linewidth}{!}{%
  \textsuperscript{1} \textit{Wichita State University, USA}\quad
  \textsuperscript{2} \textit{CCDS, Independent University, Bangladesh}\quad
  \textsuperscript{3} \textit{Amazon GenAI, USA}}
  \end{minipage}
  }

\begin{document}
    \maketitle
    {\let\thefootnote\relax
    \footnotetext{\noindent
    \begin{minipage}{\linewidth}\raggedright
    \textdagger~Equal supervision.\quad
    \textdaggerdbl~Work done outside role at Amazon.\\
    (\Letter)~Corresponding author: \texttt{mausulazad495@gmail.com}
    \end{minipage}}}
    
    
    
    \begin{abstract}
    Document packets, multiple documents concatenated into a single file, are common in government and administrative workflows, yet splitting them into their constituent documents is difficult, especially for low-resource languages. We introduce \textbf{Khondo} (Bangla for split/segment), the first benchmark for document packet splitting on Bangladeshi government forms. Unlike prior English and OCR-text-based datasets, Khondo is bilingual (Bangla--English) and vision-native; where models operate directly on page images. 
    It spans five concatenation schemes, from sequential to fully shuffled, across 14 administrative domains, with ground-truth boundaries, domain types, and page order. 
    Zero-shot evaluation of MLLMs shows they cluster pages into their source documents fairly well but struggle in restoring the original page order once shuffled. To isolate what drives this difficulty, we run two controlled analyses, varying the prompt instruction and then the packet language. 
    Both primarily affect ordering rather than clustering: (a) explicit page-order instructions are necessary but insufficient, and (b) English packets are ordered more reliably than Bangla, making page arrangement the dominant challenge and language a secondary but consistent factor.
    Khondo establishes page-order reconstruction as a key open problem in vision-based, low-resource document understanding, and provides a controlled benchmark for measuring progress toward solving it. 
    Our dataset and code is available at:
    \url{https://huggingface.co/datasets/Mausul/khondo}
    \end{abstract}


    \section{Introduction}
\label{sec:intro}

\begin{figure*}[htb]
    \centering
    \includegraphics[width=\linewidth]{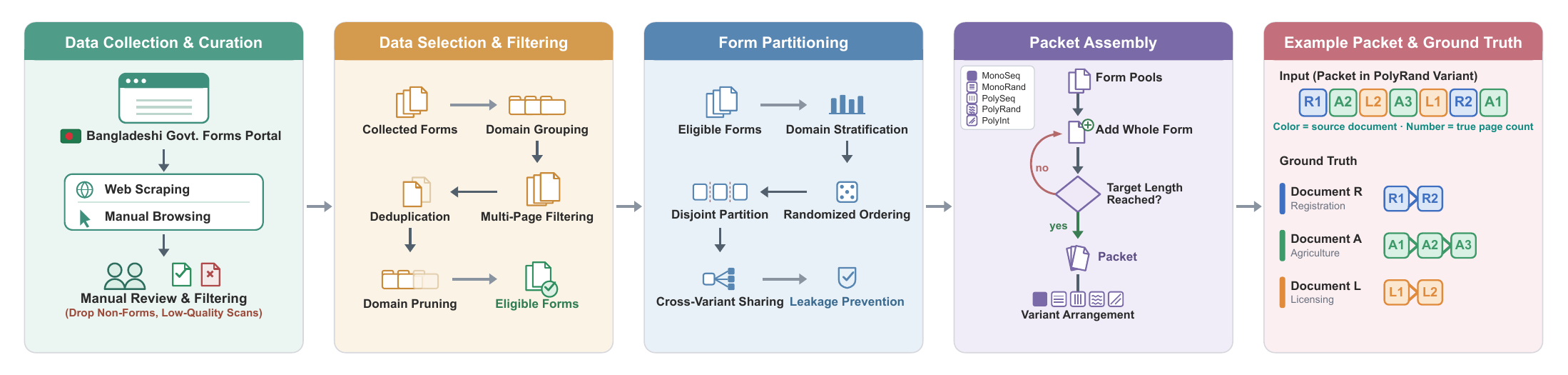}
    \caption{
        \textbf{Overview of the Khondo dataset construction pipeline.} Document packets are created by concatenating authentic Bangladeshi government forms, enabling ground-truth document membership and page order to be recorded during assembly. From left to right, the pipeline consists of: (1) document collection and curation; (2) selection of eligible multi-page forms; (3) partitioning into disjoint, leakage-free train/validation/test splits; and (4) packet generation under multiple concatenation schemes. Panel (5) illustrates the packet-splitting task: pages from several documents are concatenated and potentially shuffled. The objective is to recover both the document-level clustering and the original within-document page order. Colors denote source documents, and numbers indicate ground-truth page indices.
    }
    \label{fig:overview}
\end{figure*}

In government and administrative workflows, documents are rarely processed individually. Applications and case files are commonly scanned and archived as \emph{packets}: collections of forms and supporting materials concatenated into a single file, often with pages appearing out of their original order~\cite{un_report}. Recovering the constituent documents (identifying document boundaries, grouping pages by document, and inferring original order) is a prerequisite for downstream processing, yet this step remains largely manual.

We study this problem as \emph{document packet splitting}: clustering pages into their source documents, classifying document types, and reconstructing the original page order, even when pages are shuffled~\cite{docsplit}. Two characteristics of the setting motivate our approach. First, the documents are mostly Bangla forms, a low-resource domain where OCR remains unreliable. We therefore cast the task as \emph{vision-native}, operating directly on page images. Second, packet splitting needs structured, generative outputs that existing document models do not readily provide. Discriminative encoders cannot generate them~\cite{layoutlmv3, lilt, bros}, while generative document models typically need task-specific pretraining and are not built for prompt-based use~\cite{udop, ofa, donut}. Multimodal large language models (MLLMs), by contrast, are promptable, training-free at inference, and API-accessible~\cite{llm_cost, mllm_cost}, making them a natural fit. We therefore evaluate flagship off-the-shelf MLLMs, and fine-tune smaller open models to test the task's learnability.

Despite growing interest in document understanding, existing resources leave this setting largely unexplored. Prior work on packet splitting has focused primarily on English business documents~\cite{docsplit, tobacco}. At the same time, research on Bangla document understanding (despite Bangla being one of the world's most widely spoken languages) remains dominated by text-based tasks~\cite{zaban}, and multilingual document-understanding benchmarks typically exclude Bangla~\cite{xfund, dude, mtvqa}. To our knowledge, no Bangla benchmark exists for document packet splitting or comparable form-understanding tasks.

To address this gap, we introduce \textbf{Khondo}, a document packet-splitting benchmark centered on Bangladeshi government forms. To the best of our knowledge, it is the first such benchmark for a low-resource language. Evaluating flagship MLLMs without task-specific training, we find that clustering is comparatively robust, whereas page ordering breaks down once pages are shuffled. This contrasts with text-based packet splitting, where page ordering is largely considered a solved problem~\cite{docsplit}. Two further analyses probe this ordering gap. Explicit instructions to re-sequence pages narrows the gap but does not eliminate it, suggesting that the difficulty is intrinsic rather than a prompting artifact. The gap is also partly linguistic: identical packets are ordered more accurately in English than in Bangla, although page arrangement remains the dominant factor.

Our main contributions are as follows:
\begin{enumerate}
      \item We introduce \textbf{Khondo}, a benchmark for document packet splitting on Bangladeshi government forms, comprising 1,950 bilingual document packets across 14 administrative domains and 5 concatenation schemes, from sequential to fully shuffled.

      \item We benchmark flagship MLLMs and show that they cluster pages into documents reasonably well but struggle to recover the original page order under shuffling.

      \item We fine-tune small open-source MLLMs on Khondo, which lifts ordering well above the zero-shot baselines but still leaves a large gap on shuffled variants.

      \item We conduct two controlled studies, varying prompt phrasing and packet language, and find that both shift ordering far more than clustering.
  \end{enumerate}
    \section{Related Works}
\label{sec:related_works}

\textbf{Document Packet Splitting and Page Stream Segmentation:} Page Stream Segmentation (PSS), splitting a continuous page stream into its constituent documents, has a long history, ranging from digital-mailroom systems that jointly segment and classify documents \cite{pss_mailroom} to multimodal CNNs that fuse OCR text and page images \cite{pss_cnn, pss_cnn_journal}, clustering-based formulations \cite{pss_clustering}, and benchmarks such as TABME \cite{tabme}. More recent work applies decoder-only LLMs, which outperform encoder-based approaches \cite{llm_pss}, and multimodal transformers \cite{cosmo}, while a parallel line of research models richer inter-page structure \cite{mugat, interpage_relations, beyond_page_classif}. However, PSS assumes that pages arrive \emph{in order}; it is treated as a boundary-detection problem rather than one of order reconstruction. Document packet splitting generalizes this setting \cite{docsplit}: pages may be shuffled or interleaved, requiring the reconstruction of each document's page order in addition to document clustering and document type classification. DocSplit formalizes this problem and benchmarks MLLMs, but only on OCR-extracted text from English business documents \cite{rvlcdip}. We adopt the same formulation in a vision-native, low-resource setting (Bangladeshi government forms).

\textbf{MLLMs for Document Understanding:} MLLMs are increasingly being applied to document understanding, both in deployed agentic systems \cite{idpflow} and in document-specialized models. Because high-resolution pages produce thousands of visual tokens, processing multi-page documents is computationally expensive; recent OCR-free MLLMs address this challenge through token compression \cite{mplug_docowl2}. Multi-page comprehension is itself demanding: benchmarks for multi-page document VQA \cite{mp_docvqa} and long-context multimodal documents \cite{mmlongbench_doc} show that reasoning must span multiple pages and that even frontier models degrade as context length increases. Small MLLMs provide efficient alternatives \cite{svlm_survey}, including document-specialized variants \cite{docslm}. However, this body of work is focused on document QA, information extraction, and parsing rather than packet splitting, and is overwhelmingly English-centric.

\textbf{Bangla and Low-Resource Document Understanding:} Bangla-centric NLP covers a broad range of tasks but remains predominantly text-based. Regional Indic initiatives have produced corpora, benchmarks, and models across multiple languages \cite{indicxtreme}, while Bangla-specific models span both discriminative \cite{banglabert} and generative \cite{banglanlg, tigerllm} paradigms. However, these efforts focus on plain-text understanding rather than document understanding. For documents, Bangla resources extend only to layout analysis \cite{badlad}; to the best of our knowledge, no benchmark currently exists for Bangla key-information extraction, form understanding, or packet splitting. This gap is particularly notable given the maturity of multilingual document understanding research: popular benchmarks cover several languages \cite{xfund, mtvqa, dude} but exclude Bangla, while packet splitting has been benchmarked only on English documents \cite{docsplit}. Khondo addresses this gap by introducing the first packet-splitting benchmark for Bangla-heavy Bangladeshi government forms and the first for any low-resource language.
    \section{Preliminaries}
\label{sec:prelim}

Following DocSplit~\cite{docsplit}, we formalize \emph{document packet splitting} as recovering the individual documents concatenated into a single packet. Unlike DocSplit, which operates on OCR tokens, Khondo works directly on page images.

A \emph{packet} is a sequence of \(n\) page images \(P=\langle p_1,\dots,p_n\rangle\) formed by concatenating \(k\) source documents. Pages may appear out of order and need not form contiguous blocks. Packet splitting seeks to recover:

\begin{itemize}
  \item \textbf{Clustering}: a partition $\mathcal{D} = \{D_1, \dots, D_k\}$ of $\{1, \dots, n\}$,
    with $D_i$ the positions of document $i$'s pages.
  \item \textbf{Typing}: a map $t$ assigning each document $i$ a domain $t(i) \in \mathcal{Y}$.
  \item \textbf{Ordering}: for each document $i$, a permutation
    $\sigma_i : D_i \to \{1, \dots, |D_i|\}$ giving its original page order.
\end{itemize}

Given $P$, a method predicts $(\hat{\mathcal{D}}, \hat{t}, \{\hat{\sigma}_i\})$, approximating the ground truth $(\mathcal{D}, t, \{\sigma_i\})$. The difficulty of recovering each structure depends on how the packet was assembled.

We use five concatenation variants, each given by a domain composition and a page arrangement. \emph{Mono} variants draw all $k$ documents from one domain ($t(1) = \cdots = t(k)$); \emph{poly} variants from at least two ($|\{t(1), \dots, t(k)\}| \ge 2$).

Let \(\mathbf d_i=\langle(i,1),\dots,(i,m_i)\rangle\) denote document \(i\)'s pages in true order, with \(\oplus\), \(\mathrm{RR}\), and \(\mathrm{shuf}\)
denoting concatenation, round-robin interleaving, and uniform shuffling, respectively. The variants are therefore:
\begin{align*}
      \text{MonoSeq:}\quad  & P = \mathbf{d}_1 \oplus \cdots \oplus \mathbf{d}_k, \\
      \text{MonoRand:}\quad & P = \mathrm{shuf}(\mathbf{d}_1) \oplus \cdots \oplus \mathrm{shuf}(\mathbf{d}_k) \\
      \text{PolySeq:}\quad  & P = \mathbf{d}_1 \oplus \cdots \oplus \mathbf{d}_k, \\
      \text{PolyInt:}\quad  & P = \mathrm{RR}(\mathbf{d}_1, \dots, \mathbf{d}_k), \\
      \text{PolyRand:}\quad & P = \mathrm{shuf}(\mathbf{d}_1 \oplus \cdots \oplus \mathbf{d}_k).
\end{align*}
MonoSeq and PolySeq use the same concatenation and differ only in composition; the two randomized variants differ in the scope of the shuffle: within each document block (MonoRand) or across the whole packet (PolyRand). ($\mathrm{RR}$ emits the first page of each document, then the second...)

Arrangement alone, independent of composition, determines the page order. For documents $A = \langle A_1, A_2, A_3 \rangle$ and $B = \langle B_1, B_2 \rangle$:
\begin{align*}
    \text{ordered:}\quad & A_1,A_2,A_3,B_1,B_2, \\
    \text{within-block shuffle:}\quad & A_2,A_1,A_3,B_2,B_1, \\
    \text{round-robin:}\quad & A_1,B_1,A_2,B_2,A_3, \\
    \text{global shuffle:}\quad & A_2,B_1,A_1,A_3,B_2.
\end{align*}
The two shuffled arrangements illustrate just one possible realization, as every ordering permitted by the corresponding shuffle is equally likely. In contrast, ordered and round-robin are deterministic given the document order.

The variants isolate different sources of difficulty. Ordering is non-trivial only when shuffling disrupts page order, i.e., in MonoRand and PolyRand. Clustering is most difficult when documents are interleaved (PolyInt, PolyRand) or when all documents share the same domain, thereby eliminating type cues (MonoSeq, MonoRand).

    \section{Dataset}
\label{sec:dataset}

\subsection{Data Collection and Packet Assembly}
\cref{fig:overview} summarizes the construction pipeline of Khondo. The dataset is built from publicly available Bangladeshi government forms collected from the national e-forms portal\footnote{\url{https://forms.portal.gov.bd/}} across 19 administrative domains. Each file was reviewed manually to discard non-forms and low-quality scans. Because Bangladeshi administrative workflows routinely mix Bangla and English (many official forms, particularly in finance, commerce, and registration, are partly or entirely in English), real document packets are inherently bilingual. See \ref{subsec:lang_comp} for a breakdown of form languages.

Before packet assembly, we partition the 423 forms at the form level, using a 55:20:25 train:validation:test split, stratified by domain following DocSplit~\cite{docsplit}. The partitions are disjoint, so no source form appears in more than one split, preventing content-level leakage. We then assemble the five variants within each split. 

For each split and variant, we assemble a packet by sampling a target length of 5--20 pages, a range every evaluated model can process reliably. We append whole forms (each used at most once) to the packet till the target length is reached or no further form fits within target range. We sample domains uniformly across packets, so all 14 stay within ${\sim}1.6$ percentage points of the 7.1\% uniform line and create a near-uniform distribution where no high-frequency domain dominates the overall performance (See \cref{fig:domain-dist} in \cref{subsec:domain_dist}). Document membership, domain type, and page order are recorded as ground truth at assembly time.

\subsection{Dataset Statistics}
Khondo comprises \textbf{1{,}950 packets} (390 per variant), totaling \textbf{28{,}513 pages}, drawn from 423 eligible source forms across 14 domains. Packets span 5--20 pages (mean: 14.6) and contain 1--9 sub-documents (mean: 3.8); \cref{tab:dataset-stats} summarizes per-variant statistics. Following an evaluation-first design, the majority of packets are allocated to the test split (210 per variant) to enable statistically robust evaluation.

The benchmark is bilingual: at the page level, it comprises 67\% Bangla, 26\% English, and 5\% mixed Bangla--English pages; with 72\% of all pages containing Bangla script. Per-page language labels are assigned using a dual-MLLM pipeline with a frontier-model tiebreaker, followed by blind dual-human review of ambiguous cases, achieving an inter-annotator agreement of Cohen's $\kappa = 0.90$. See \ref{subsec:lang_label} for more details.

\begin{table}[t]
    \centering
    \small
    \setlength{\tabcolsep}{5pt}
    \caption{Statistics for 5 Khondo concatenation variants. \emph{Mean Len.}\ is mean pages per packet, \emph{Mean Sub-Docs} is mean number of source documents per packet. Packets are sampled uniformly across domains, yielding a domain-balanced benchmark in which each of the 14 domains accounts for 5.6--8.8\% of sub-documents.}
    \label{tab:dataset-stats}

    \resizebox{0.9\columnwidth}{!}{
        \begin{tabular}{ccrcc}
            \toprule
            \textbf{Variant}  & \textbf{Packets} & \textbf{Pages} & \textbf{Mean Len.} & \textbf{Mean Sub-Docs} \\
            \midrule
            MonoSeq  &   390 &  5{,}472 & 14.0 & 3.71 \\
            MonoRand &   390 &  5{,}329 & 13.7 & 3.50 \\
            PolySeq  &   390 &  6{,}023 & 15.4 & 4.12 \\
            PolyInt  &   390 &  5{,}809 & 14.9 & 3.92 \\
            PolyRand &   390 &  5{,}880 & 15.1 & 3.95 \\
            \midrule
            All      & 1{,}950 & 28{,}513 & 14.6 & 3.84 \\
            \bottomrule
        \end{tabular}
    }
\end{table}
    \section{Experiment Details}
\label{sec:experiments}

\subsection{Inference Setup and Evaluation Protocol}

\textbf{Inference:} Each packet $P=\langle p_1,\dots,p_n\rangle$ is presented to a model $M_\theta$ as a sequence of $n$ page images with a single zero-shot prompt $q$ instructing it to group pages into their source documents, reconstruct each document's page order, and classify its domain. The prompt specifies a JSON output schema (a list of sub-documents, each with a domain label and ordered page list) for reliable extraction: 
\[ M_\theta(P,q)=(\hat{\mathcal{D}},\hat{t},\{\hat{\sigma}_i\}), \] 
Structurally invalid outputs are retried up to $k=3$ times.

\textbf{Metrics:} \emph{Clustering} compares the predicted and ground-truth page partitions using the mean of the V-measure~\cite{v_measure} and Rand index~\cite{ri_score}, both invariant to sub-document labels: 
\begin{equation} 
    S_{\mathrm{clu}}=\tfrac12\mathrm{V}(\mathcal{D},\hat{\mathcal{D}})+\tfrac12\mathrm{RI}(\mathcal{D},\hat{\mathcal{D}}) 
\end{equation} 
Within each predicted multi-page sub-document, \emph{Ordering} compares the predicted page order with the ground-truth order using Kendall's $\tau_a$ over its $m$ pages~\cite{tau}: 
\begin{equation} 
    \tau_a=\frac{n_c-n_d}{\binom{m}{2}} 
\end{equation} 
where $n_c$ and $n_d$ denote the numbers of concordant and discordant page pairs, respectively ($\tau_a=1$ indicates a perfect ordering). $S_{\mathrm{ord}}$ is the mean $\tau_a$ over all predicted multi-page sub-documents (defined as $1$ if a packet contains none).

The overall packet score equally weights the two components, $S_{\mathrm{pkt}}=\tfrac12 S_{\mathrm{clu}}+\tfrac12 S_{\mathrm{ord}}$, following DocSplit. Like DocSplit, $S_{\mathrm{pkt}}$ evaluates partition \emph{structure} rather than labels. We therefore report \emph{typing} separately as page accuracy, $\mathrm{PgAcc}=\frac{1}{n}\sum_{j=1}^{n}\mathds{1}\!\left[\hat{t}(j)=t(j)\right]$, alongside DocSplit's page, page+split, and page+split+order accuracies. All metrics are averaged over the test packets.

\subsection{Training-Free Zero-Shot Inference}
\label{subsec:zeroshot}

We evaluate five flagship MLLMs spanning proprietary and open-weight ecosystems: the proprietary \textbf{Gemini-3.5-Flash}, \textbf{GPT-5.4}, and \textbf{Qwen~3.6 Plus}, and the open-weight \textbf{GLM-4.6V} and \textbf{Kimi~K2.5}, all accessed through OpenRouter\footnote{\url{https://openrouter.ai/}}. Their open-ended, instruction-following inference matches the target setting: splitting a packet directly from page images, without OCR or task-specific training. Each packet is submitted in a single request containing all $n$ page images, using the order-aware (\emph{OAw}) prompt. The prompt restricts domain labels to the benchmark taxonomy, pairing each with a one-line description. We report results on the test split (210 packets per variant) and evaluate all models with $temperature=1.0$ and reasoning disabled.

\subsection{Supervised Fine-Tuning}
\label{subsec:sft}

In addition to flagship MLLMs, we benchmark \emph{small}, \emph{open}-weight models: \textbf{Qwen3-VL-4B}, \textbf{InternVL3.5-4B}, and \textbf{Gemma~4~E2B}. Each is adapted with QLoRA (4-bit, rank~16) using its native non-reasoning chat template to match the zero-shot setting. To ensure a fair comparison, we cap the per-page image resolution at approximately $512\times512$, equalizing the per-page token budget while keeping multi-page packets within GPU memory. We train a single adapter per backbone on the pooled training set from all five variants (150 packets per variant), holding out 30 packets per variant for model selection, on a single NVIDIA~L40S. Evaluation uses the same test split as the zero-shot models.

\subsection{Prompt Sensitivity}
\label{subsec:prompt_sense}
We use an \textbf{order-aware (OAw)} prompt (\ref{subsec:prompt}) for benchmarking, which explicitly states that pages within a document may appear out of order and must be re-sequenced. To gauge how much of a model's ordering performance is anchored around this instruction, we re-score the \emph{identical} packets with an \textbf{order-agnostic (OAg)} prompt (\ref{subsec:prompt}) that contains no instruction regarding page order. Because the same packets are scored under both prompts, the comparison is paired: $\Delta S_{\mathrm{ord}} = \text{OAw}-\text{OAg}$ per `model x order variant'. We evaluate the three strongest zero-shot (training-free) models---\textbf{Gemini-3.5-Flash}, \textbf{Qwen~3.6 Plus}, and \textbf{GPT-5.4}---using samples of $100$ packets per variant. We measure prompt sensitivity in training-free setup, to keep models un-exposed to our training data and subsequently to control influence of learned representations so that we can properly isolate effects of prompt alteration

\subsection{Cross-Lingual Effects}
\label{subsec:xling}
Khondo's packets are bilingual and Bangla-dominant, reflecting the underlying corpus of Bangladeshi government forms (\ref{subsec:lang_comp}). This raises a key question about our central finding: is the page-ordering bottleneck driven by the difficulty of reading Bangla script, or by page arrangement independent of language? To disentangle these factors, we construct controlled monolingual packets---\textbf{Bangla}-only and \textbf{English}-only---by assembling each of the five variants from single-language source forms while holding the packet-size distribution fixed. The two conditions therefore differ only in language (100 test packets per language per variant). We evaluate the same three models used in the prompt-sensitivity study across all five variants and report the per-variant differences, $\Delta S_{\mathrm{ord}}=\mathrm{En}-\mathrm{Bn}$, for ordering, clustering, and page accuracy. As in the prompt-sensitivity study, we evaluate only training-free models so that any observed differences reflect language rather than representations learned from our training data.

    \section{Results and Discussion}
\label{sec: results}


\begin{table}[t]
    \centering\small
    \setlength{\tabcolsep}{5pt}
    \caption{         
        Zero-shot packet-splitting performance on the Khondo test split (210 packets per variant) using the order-aware (OAw) prompt.  $S_{\mathrm{clu}}$ and $S_{\mathrm{ord}}$ denote the clustering and ordering scores, respectively, and $S_{\mathrm{pkt}}=\tfrac{1}{2}(S_{\mathrm{clu}}+S_{\mathrm{ord}})$ is their average. PgAcc denotes page-level typing accuracy. The lower block reports results on the shuffled variants. \textbf{Bold} indicates the best-performing model for each metric within each variant. Across the shuffled variants, $S_{\mathrm{ord}}$ drops sharply while $S_{\mathrm{clu}}$ remains relatively stable, highlighting page ordering as the primary bottleneck. Higher ($\uparrow$) indicates better performance for all metrics.
    }
    \label{tab:zeroshot}
    \resizebox{0.9\columnwidth}{!}{
        \begin{tabular}{ll cccc}
            \toprule
            \textbf{Variant} & \textbf{Model} & \textbf{$S_{\mathrm{clu}}$} & \textbf{$S_{\mathrm{ord}}$} & \textbf{$S_{\mathrm{pkt}}$} & PgAcc \\
            \midrule
            \multirow{5}{*}{MonoSeq}
            & Kimi K2.5         & 0.774 & 0.738 & 0.756 & 0.516 \\
            & GLM-4.6V          & 0.706 & 0.839 & 0.773 & 0.318 \\
            & GPT-5.4           & 0.819 & 0.840 & 0.829 & 0.529 \\
            & Qwen 3.6 Plus     & \textbf{0.867} & 0.927 & \textbf{0.897} & \textbf{0.579} \\
            & Gemini-3.5-Flash  & 0.847 & \textbf{0.947} & \textbf{0.897} & 0.501 \\
            \midrule
            \multirow{5}{*}{PolySeq}
            & Kimi K2.5         & 0.836 & 0.711 & 0.774 & 0.406 \\
            & GLM-4.6V          & 0.748 & 0.909 & 0.829 & 0.172 \\
            & GPT-5.4           & 0.854 & 0.806 & 0.830 & \textbf{0.513} \\
            & Qwen 3.6 Plus     & \textbf{0.899} & 0.944 & \textbf{0.922} & 0.511 \\
            & Gemini-3.5-Flash  & 0.864 & \textbf{0.953} & 0.909 & 0.510 \\
            \midrule[1.2pt]
            \multirow{5}{*}{MonoRand}
            & Kimi K2.5         & 0.749 & 0.092 & 0.421 & 0.506 \\
            & GLM-4.6V          & 0.690 & 0.225 & 0.457 & 0.292 \\
            & GPT-5.4           & 0.815 & 0.609 & 0.712 & 0.567 \\
            & Qwen 3.6 Plus     & \textbf{0.852} & 0.572 & 0.712 & \textbf{0.569} \\
            & Gemini-3.5-Flash  & 0.845 & \textbf{0.815} & \textbf{0.830} & 0.529 \\
            \midrule
            \multirow{5}{*}{PolyInt}
            & Kimi K2.5         & 0.529 & 0.070 & 0.300 & 0.281 \\
            & GLM-4.6V          & 0.625 & 0.130 & 0.378 & 0.164 \\
            & GPT-5.4           & 0.706 & 0.469 & 0.587 & 0.415 \\
            & Qwen 3.6 Plus     & \textbf{0.825} & 0.599 & 0.712 & \textbf{0.502} \\
            & Gemini-3.5-Flash  & 0.782 & \textbf{0.676} & \textbf{0.729} & 0.461 \\
            \midrule
            \multirow{5}{*}{PolyRand}
            & Kimi K2.5         & 0.547 & 0.051 & 0.299 & 0.315 \\
            & GLM-4.6V          & 0.642 & 0.151 & 0.397 & 0.167 \\
            & GPT-5.4           & 0.706 & 0.521 & 0.614 & 0.422 \\
            & Qwen 3.6 Plus     & \textbf{0.833} & 0.601 & 0.717 & \textbf{0.497} \\
            & Gemini-3.5-Flash  & 0.779 & \textbf{0.675} & \textbf{0.727} & 0.475 \\
            \bottomrule
        \end{tabular}
    }
\end{table}

\textbf{Under training-free zero-shot inference, all five flagship MLLMs group pages into sub-documents far more accurately than they reconstruct page order on the shuffled variants.} On the sequential variants, ordering appears strong ($S_{\mathrm{ord}}$ up to $0.95$), but only because pages are already in their original order; the shuffled variants remove this advantage. Clustering degrades much less than ordering, with the stronger models maintaining $S_{\mathrm{clu}}$ between $0.55$ and $0.85$, whereas ordering ranges from $0.82$ for the best model to $0.05$ for the weakest, which largely preserves the input order. This is the opposite of DocSplit, where ordering exceeds $0.97$ on contiguous packets and clustering is the more challenging task. On Khondo, \emph{the primary challenge is reconstructing page order after shuffling, not determining which pages belong together.} Gemini-3.5-Flash performs best overall, with Qwen 3.6 Plus a clear second, both outperforming GPT-5.4, while GLM-4.6V and Kimi K2.5 trail well behind. The two leaders perform similarly on the sequential variants and differ mainly on shuffled ordering, where Gemini achieves the highest ordering score ($S_{\mathrm{ord}}=0.68$--$0.82$ vs.\ $0.57$--$0.60$ for Qwen). See \Cref{tab:zeroshot} for the full results.

\begin{table}[t]
    \centering\small
    \setlength{\tabcolsep}{4pt}
    \caption{
        Supervised fine-tuning results using the order-aware (OAw) prompt on the Khondo test split (210 packets per variant). Baseline model performance is denoted by \emph{base} and \emph{FT} denotes performance after fine-tuning. $S_{\mathrm{clu}}$, $S_{\mathrm{ord}}$, $S_{\mathrm{pkt}}$, and PgAcc are defined as in \Cref{tab:zeroshot}. The lower block reports results on the shuffled variants. \textbf{Bold} indicates the best-performing fine-tuned model for each metric within each variant. PolyInt is independently sampled using full-page shuffling, yielding a distinct set of packets from PolyRand while maintaining a matched difficulty distribution.
    }
    \label{tab:sft}
    \resizebox{0.9\columnwidth}{!}{
        \begin{tabular}{lll cccc}
            \toprule
            \textbf{Variant} & \textbf{Model} & & \textbf{$S_{\mathrm{clu}}$} & $S_{\mathrm{ord}}$ & \textbf{$S_{\mathrm{pkt}}$} & PgAcc \\
            \midrule
            
            \multirow{6}{*}{MonoSeq} & \multirow{2}{*}{Gemma~4~E2B} & Base & 0.554 & 0.490 & 0.522 & 0.193 \\
            & & FT & 0.585 & 0.625 & 0.605 & 0.451 \\
            \cmidrule(l){2-7}
            & \multirow{2}{*}{InternVL3.5-4B} & Base & 0.361 & 0.446 & 0.404 & 0.284 \\
            & & FT & \textbf{0.881} & 0.731 & \textbf{0.806} & 0.452 \\
            \cmidrule(l){2-7}
            & \multirow{2}{*}{Qwen3-VL-4B} & Base & 0.591 & 0.612 & 0.601 & 0.297 \\
            & & FT & 0.843 & \textbf{0.761} & 0.802 & \textbf{0.566} \\
            \midrule
            
            \multirow{6}{*}{PolySeq} & \multirow{2}{*}{Gemma~4~E2B} & Base & 0.592 & 0.581 & 0.586 & 0.115 \\
            & & FT & 0.659 & 0.640 & 0.650 & 0.202 \\
            \cmidrule(l){2-7}
            & \multirow{2}{*}{InternVL3.5-4B} & Base & 0.404 & 0.465 & 0.434 & 0.135 \\
            & & FT & 0.937 & 0.855 & 0.896 & \textbf{0.286} \\
            \cmidrule(l){2-7}
            & \multirow{2}{*}{Qwen3-VL-4B} & Base & 0.582 & 0.639 & 0.610 & 0.169 \\
            & & FT & \textbf{0.945} & \textbf{0.866} & \textbf{0.906} & 0.283 \\
            \midrule[1.2pt]
            \multirow{6}{*}{MonoRand} & \multirow{2}{*}{Gemma~4~E2B} & Base & 0.561 & 0.009 & 0.285 & 0.128 \\
            & & FT & 0.563 & $-$0.036 & 0.264 & 0.479 \\
            \cmidrule(l){2-7}
            & \multirow{2}{*}{InternVL3.5-4B} & Base & 0.307 & 0.023 & 0.165 & 0.245 \\
            & & FT & 0.819 & 0.190 & 0.505 & 0.448 \\
            \cmidrule(l){2-7}
            & \multirow{2}{*}{Qwen3-VL-4B} & Base & 0.515 & 0.023 & 0.269 & 0.271 \\
            & & FT & \textbf{0.840} & \textbf{0.316} & \textbf{0.578} & \textbf{0.528} \\
            \midrule
            
            \multirow{6}{*}{PolyInt} & \multirow{2}{*}{Gemma~4~E2B} & Base & 0.376 & $-$0.005 & 0.186 & 0.105 \\
            & & FT & 0.420 & 0.043 & 0.231 & 0.219 \\
            \cmidrule(l){2-7}
            & \multirow{2}{*}{InternVL3.5-4B} & Base & 0.280 & 0.005 & 0.142 & 0.115 \\
            & & FT & 0.696 & 0.156 & 0.426 & 0.249 \\
            \cmidrule(l){2-7}
            & \multirow{2}{*}{Qwen3-VL-4B} & Base & 0.300 & $-$0.009 & 0.146 & 0.146 \\
            & & FT & \textbf{0.788} & \textbf{0.333} & \textbf{0.561} & \textbf{0.270} \\
            \midrule
            
            \multirow{6}{*}{PolyRand} & \multirow{2}{*}{Gemma~4~E2B} & Base & 0.389 & 0.103 & 0.246 & 0.096 \\
            & & FT & 0.400 & 0.049 & 0.224 & 0.230 \\
            \cmidrule(l){2-7}
            & \multirow{2}{*}{InternVL3.5-4B} & Base & 0.268 & 0.027 & 0.148 & 0.132 \\
            & & FT & 0.695 & 0.189 & 0.442 & 0.246 \\
            \cmidrule(l){2-7}
            & \multirow{2}{*}{Qwen3-VL-4B} & Base & 0.261 & 0.033 & 0.147 & 0.127 \\
            & & FT & \textbf{0.797} & \textbf{0.321} & \textbf{0.559} & \textbf{0.276} \\
            \bottomrule
         \end{tabular}
    }
\end{table}

\textbf{Fine-tuning small open-source MLLMs does not change the overall picture: clustering improves, but ordering remains the bottleneck.} We fine-tune three 4B-class MLLMs (Qwen3-VL-4B, InternVL3.5-4B, and Gemma4-E2B) and compare each with its pre-fine-tuning base model (\Cref{tab:sft}). After training, the stronger models achieve good ordering on the sequential variants ($S_{\mathrm{ord}}=0.73$--$0.87$), but ordering remains poor on the shuffled variants: the best small MLLM, Qwen3-VL-4B, reaches only $S_{\mathrm{ord}}\approx0.32$, despite maintaining clustering around $0.80$. The base models barely recover page order under shuffling ($S_{\mathrm{ord}}\approx0$), so fine-tuning substantially improves ordering without eliminating the gap. Thus, page ordering remains the primary bottleneck for small open-source MLLMs, just as it does for the flagship models. The gains are also uneven across backbones: Qwen3-VL-4B and InternVL3.5-4B improve markedly (PolyRand $S_{\mathrm{pkt}}$ from $0.15$ to $0.56$ and from $0.44$ to $0.56$, respectively), whereas Gemma4-E2B, the smallest model at roughly 2B effective parameters, declines slightly ($0.25$ to $0.22$). The clustering--labeling gap observed in the zero-shot models also persists after fine-tuning: \emph{clustering reaches up to $0.95$ (PolySeq), while page accuracy never exceeds $0.57$.}

\begin{figure}[ht]
    \centering
    \includegraphics[width=\linewidth]{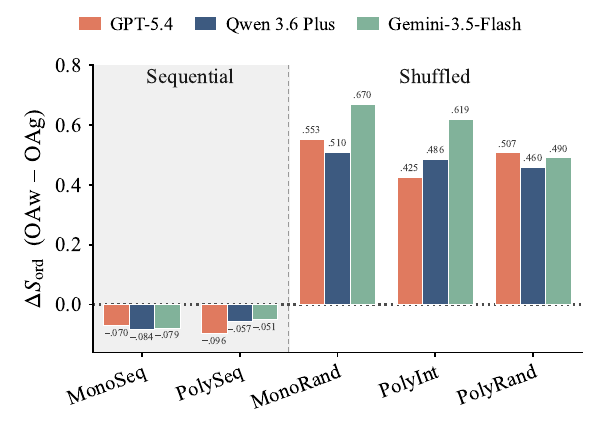}
    \caption{ 
        Change in ordering score, $\Delta S_{\mathrm{ord}} = S_{\mathrm{ord}}^{\mathrm{OAw}} - S_{\mathrm{ord}}^{\mathrm{OAg}}$, for each model and packet variant. On the shuffled variants, the order-aware (OAw) prompt recovers $0.43$--$0.67$ of the ordering performance lost by the order-agnostic (OAg) prompt, which otherwise tends to reproduce the presented page order. On the sequential variants, where the input pages are already correctly ordered, the explicit ordering cue slightly reduces performance ($\Delta S_{\mathrm{ord}}=-0.05$ to $-0.10$).
    }
    \label{fig:prompt-ord}
\end{figure}

\textbf{Ordering is highly prompt-sensitive, but a substantial deficit remains even with an order-aware prompt.} The bottleneck in the previous experiments was measured using a single order-aware (OAw) prompt, raising the question of whether it stems from the prompt rather than the task itself. We therefore re-evaluate the same packets on the three frontier models using an order-agnostic (OAg) prompt that omits instructions to reconstruct page order (\Cref{fig:prompt-ord}). Prompt wording has a large effect, particularly on the shuffled variants (MonoRand, PolyInt, and PolyRand). On these variants, order awareness increases $S_{\mathrm{ord}}$ from $0.03$--$0.19$ to $0.48$--$0.79$ (a gain of $0.43$--$0.67$). On the sequential variants (MonoSeq and PolySeq), where pages already appear in order, it instead incurs a small penalty ($0.05$--$0.10$) by introducing unnecessary reordering. The poor performance under the order-agnostic prompt is largely mechanical: the models simply preserve the input page order in $96$--$100\%$ of cases, while the order-aware prompt reduces this "copy-the-input" behavior to roughly one-third of predictions. Order-specific instructions are therefore necessary, but not sufficient. Even with explicit prompting, ordering on the shuffled variants remains well below that on the sequential variants, indicating a genuine difficulty with page-order reconstruction rather than prompt wording. In contrast, prompting has little effect on clustering ($|\Delta|<0.09$; see \Cref{subsec:prompt_sense}), confirming that its influence—and its limitations—are largely confined to page ordering.

\begin{figure*}[t]
  \centering
  \includegraphics[width=0.9\linewidth]{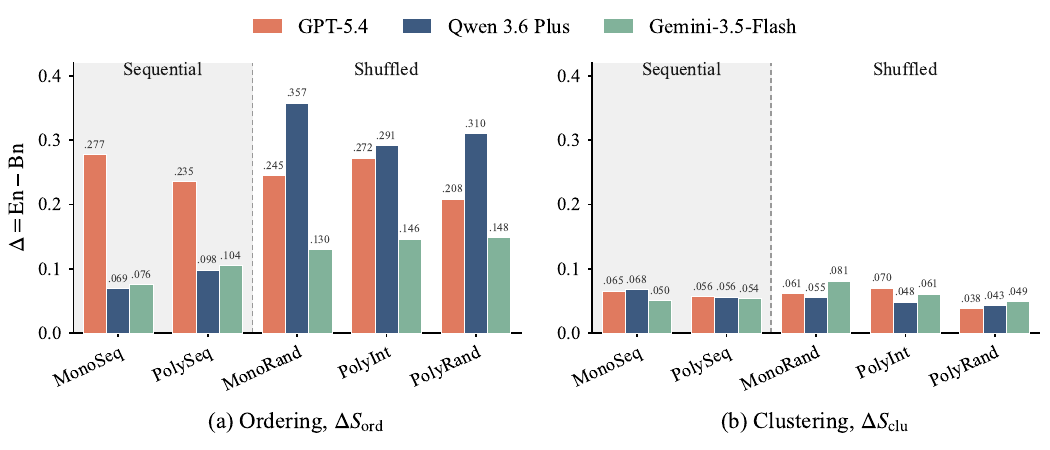}
  \caption{
      Cross-lingual performance gap, $\Delta = \mathrm{En}-\mathrm{Bn}$, on size-matched monolingual packets for each model and packet variant (shared scale). \textbf{(a)} Ordering: English packets are ordered substantially more accurately than Bangla packets, particularly on the shuffled variants, with the magnitude of the gap varying across models. \textbf{(b)} Clustering: the language effect is considerably smaller ($\Delta \approx 0.04$--$0.08$), although it remains consistently positive. Full results are reported in \Cref{tab:xling}.
  }
  \label{fig:xling}
\end{figure*}

\textbf{Part of the ordering difficulty is linguistic: MLLMs reconstruct page order substantially better for English than Bangla packets, while clustering is much less affected.} To isolate language effects in Khondo's bilingual setting, we evaluate size-matched monolingual Bangla and English packets on the three frontier models (\Cref{fig:xling}). All three achieve higher ordering scores on English packets, with the largest gaps appearing on the shuffled variants ($\Delta S_{\mathrm{ord}}=0.13$--$0.36$), although the magnitude varies by model. Clustering changes far less ($\Delta S_{\mathrm{clu}}\approx0.04$--$0.08$), indicating that language primarily affects page-order reconstruction rather than grouping. Because Khondo is naturally Bangla-dominant---reflecting the underlying corpus of Bangladeshi government forms rather than an imposed design choice---this gap lowers performance on realistic, low-resource document distributions. We deliberately preserved this natural language skew instead of balancing the corpus to evaluate models under the conditions encountered in real-world Bangladeshi document packet splitting.

    \section{Conclusion}
\label{sec:conclusion}

We introduce \textbf{Khondo}, the first multimodal benchmark for document packet splitting on Bangladeshi government forms and, to the best of our knowledge, the first such benchmark for a low-resource language. Khondo comprises document packets across five concatenation variants, ranging from sequential to fully shuffled page arrangements. Each prediction is evaluated along three axes: clustering pages into documents, ordering pages within documents, and assigning page-level document types.

We evaluate five flagship MLLMs in the zero-shot setting and find that clustering remains relatively robust under page shuffling, whereas ordering degrades substantially. 
Across all models, the sequential-to-shuffled drop in ordering far exceeds that in clustering, identifying page ordering as the primary bottleneck on Bangla-centric document packets. 
Fine-tuning small MLLMs on Khondo improves ordering over their baselines, showing that the task is learnable, although a substantial gap remains on the hardest shuffled variants. 
Explicit page-order instructions further improve ordering but do not close this gap, suggesting that the challenge is intrinsic rather than a consequence of prompt design.
Finally, we show that the task is partly linguistic: under identical page arrangements, English packets are ordered more accurately than Bangla packets by $\Delta S_{\mathrm{ord}}=0.13$--$0.36$ on the shuffled variants. Because Khondo preserves the naturally Bangla-dominant distribution of real Bangladeshi government forms, this language gap represents a realistic deployment challenge.

    \section{Limitations and Future Work}
\label{sec:limit_and_future}

Possible future directions extending Khondo include:

(a) \textbf{Packet construction}: Khondo follows the five concatenation variants introduced by DocSplit. Alternative shuffling strategies could isolate different stressors. For example, globally shuffling pages in a MonoRand packet would primarily stress document clustering, whereas shuffling pages only within individual documents in a PolyRand packet would better isolate the page-ordering problem.

(b) \textbf{Model adaptation}: we evaluated flagship MLLMs in a training-free setting and investigates lightweight adaptation through QLoRA fine-tuning of 4B-class multimodal backbones. Future work could examine larger multimodal backbones, alternative parameter-efficient fine-tuning methods, and reinforcement learning methods to better understand the remaining ordering gap.

(c) \textbf{Language and domain coverage}: Khondo consists of Bangla and English Bangladeshi government forms. Addition of South Asian languages, scripts, document types, and administrative regions would enable broader evaluation of multilingual document packet splitting.
    
    
    {
        \small
        \bibliographystyle{ieeenat_fullname}
        \bibliography{main}
    }
    
    \clearpage
    \appendix
    \onecolumn
\section{Appendix/Supplementary Materials}


\subsection{Inference Prompt Details}

Prompts used for the packet splitting experiments are detailed in this section.

\label{subsec:prompt}

\begin{figure*}[htb]
    \begin{cardbox}{Domain Label Set: \texttt{\{label\_block\}}}
      \scriptsize
      \setlength{\extrarowheight}{2pt}
      \begin{tabularx}{\linewidth}{@{}>{\ttfamily}l@{\hspace{6pt}}>{\raggedright\arraybackslash}X@{}}
          agriculture  & agriculture, farming, fertilizer, and seed forms \\ 
          \hline
          city\_corp   & city corporation (urban municipal) administrative forms \\ 
          \hline
          commerce     & Industry \& Commerce ministry forms --- industrial registration, trade, business permits (NOT banking) \\ 
          \hline
          complaints   & grievance/complaint mechanism forms --- any subject; identified by being a complaint, not by topic \\ 
          \hline
          education    & education ministry forms --- schools, scholarships, exams, admission \\ 
          \hline
          employment   & employment forms --- job applications, HR, leave, retirement, salary records \\ 
          \hline
          environment  & environment \& forest ministry forms \\ 
          \hline
          finance      & finance \& banking forms --- loans, bank accounts, financial assistance, tax \\ 
          \hline
          health       & health ministry forms --- medical, hospital, public health, incl.\ health-sector licensing \\ 
          \hline
          legal        & court \& law-enforcement forms --- case filings, summons, court registration, police verification \\ 
          \hline
          licensing    & general licensing \& permit forms (trade, professional, recreational); prefer the domain-specific category when one fits \\ 
          \hline
          local\_govt  & local government (Union\slash Upazila\slash Pourashava) forms \\ 
          \hline
          registration & civil/government registration --- birth, death, marriage, voter, NID, business registration \\ 
          \hline
          telecom      & information, postal, \& telecommunications ministry forms \\
      \end{tabularx}
    \end{cardbox}
    \caption{Administrative domain labels of the government forms in Khondo. For use in the inference prompts }
    \label{box:domain_dist}
\end{figure*}
\begin{figure*}[htb]
\begin{promptbox}{Order-Agnostic (OAg) Prompt}
You are given {n_pages} pages from a document packet. The packet is a concatenation of multiple sub-documents (forms).

Your task is to identify the sub-document boundaries. For each page, decide which sub-document it belongs to. Group consecutive pages from the same sub-document together.

Return ONLY a single JSON object with this exact schema:

{
    "subdocuments": [
      {
        "doc_type_id": "<one of the valid domain slugs below>",
        "local_doc_id": "<doc_type_id>-<2-digit ordinal starting at 01>",
        "page_ordinals": [<list of 1-based page numbers from this packet that belong to this sub-document>]
      },
      ...
    ]
  }

Constraints:
- Every page from 1 to {n_pages} MUST appear in exactly one subdocument's page_ordinals (no gaps, no duplicates).
- page_ordinals within each subdocument must be in the original page order.
- local_doc_id must follow the format "<doc_type_id>-<NN>" with a zero-padded 2-digit ordinal (e.g., "employment-01", "employment-02"). Number subdocuments of the same doc_type_id sequentially starting at 01.
- doc_type_id must be one of the values in the list below. If you genuinely cannot tell, use "unknown".

Valid doc_type_id values:
{label_block}

Respond with only the JSON object -- no explanation, no markdown fencing.
\end{promptbox}
\caption{Order Agnostic Prompt (OAg) used for the packet splitting experiments using MLLMs. Compared to the Order-Aware Prompt in \Cref{prompt:oaw}, the OAg prompt does not include instructions regarding page ordering.}
\label{prompt:oag}
\end{figure*}
\begin{figure*}[htb]
\begin{promptbox}{Order-Aware (OAw) Prompt}
You are a document classification expert. You analyze the pages of a document packet, group them into the individual forms they belong to, reconstruct each form's page order, and classify each form by its DOMAIN. Your output must be valid JSON in the requested format.

You are given {n_pages} page images. They are shown as Page 1, Page 2, ... Page {n_pages} in the order they appear in the packet -- |this is PRESENTATION order only|. The packet mixes multiple forms from various domains, and a single form may span several pages.

## Splitting Guidance
When deciding whether two pages belong to the same form:
- Content continuity: continuing fields, sections, paragraphs, or numbering.
- Visual/formatting consistency: shared layout, headers, footers, letterhead, styling.
- Logical completion: a form has a beginning, body, and end (title/cover -> fields -> signatures/closing).
- Form boundaries: a new title page, cover sheet, or clearly different subject marks a new form.
- Subject similarity: pages about the same matter likely belong together.
- |Shuffled Pages: pages in the packet MAY be shuffled out of order. Pages of one form may be non-adjacent and out of their original sequence.|
- Multiple forms of the same domain: a packet may contain several distinct forms from the same domain -- keep them as separate forms; do not merge them.

## Classification Process
1. Assign each page to its form, grouping by the cues above. Pages of one form need NOT be adjacent.
2. |Within each form, reconstruct the original page order|: begin from the page that opens the form, then iterate over its remaining pages to find the best sequential continuation -- using printed page numbers, section/heading continuity, text or tables that carry across pages, form-field flow, and closings/signatures that mark a final page.
3. Classify each form by its DOMAIN, using ONLY the domain values listed below. Do not create, invent, or use any value not in the list.

## Output
Return ONLY a single JSON object with this exact schema:

  {
    "subdocuments": [
      {
        "doc_type_id": "<a domain slug from the list below>",
        "local_doc_id": "<doc_type_id>-<2-digit ordinal starting at 01>",
        "page_ordinals": [<this form's packet page numbers, in reconstructed original order>]
      },
      ...
    ]
  }

Constraints:
- Every page from 1 to {n_pages} MUST appear in exactly one form's page_ordinals (no gaps, no duplicates).
- page_ordinals are the page labels in THIS packet (1 = first page shown). |They reflect presentation order ONLY, not original form order. List each form's pages in reconstructed original order, earliest page first.| A form's pages may be non-contiguous in the packet (e.g. [2, 5, 9]).
- local_doc_id must follow the format "<doc_type_id>-<NN>" with a zero-padded 2-digit ordinal. Number subdocuments of the same doc_type_id sequentially starting at 01.
- doc_type_id must be one of the values in the list below. If you genuinely cannot tell, use "unknown".

Valid doc_type_id (domain) values:
{label_block}

Respond with only the JSON object -- no explanation, no markdown fencing.
\end{promptbox}
\caption{Order-Aware Prompt (OAw) used for the packet splitting experiments using MLLMs. Compared to the Order-Agnostic Prompt (OAg) in \Cref{prompt:oag}, the OAw prompt explicitly instructs the models about page-ordering, which we find empirically to improve packet splitting performance. \textbf{\texttt{Bold}} text marks sentences where the order-aware (OAw) and order-agnostic (OAg) prompts differ.}
\label{prompt:oaw}
\end{figure*}

\twocolumn

\subsection{Additional Dataset Statistics}
\begin{figure}[htb]
    \centering
    \includegraphics[width=0.9\linewidth]{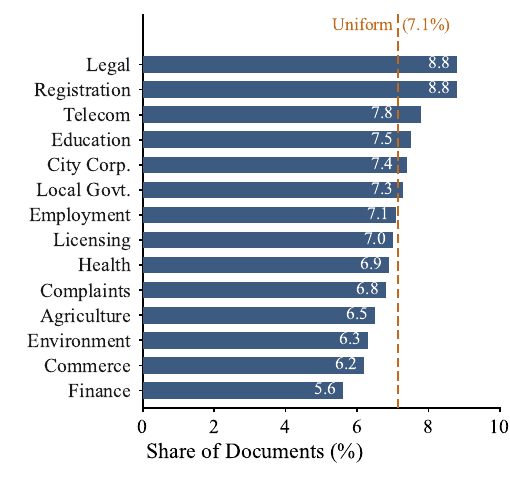}
    \caption{Domain distribution of Khondo. Each bar gives a domain's share of the 7{,}484 constituent documents across all 1{,}950 packets. All 14 domains lie within ${\sim}1.6$ percentage points of the 7.1\% uniform share (dashed line), a near-uniform distribution in which no single domain dominates.}
    \label{fig:domain-dist}
\end{figure}

In this section we report the language composition (\Cref{tab:lang-dist}) and the administrative domain distribution (\Cref{fig:domain-dist}) of the forms that make up Khondo.

\label{subsec:lang_comp}
\begin{table}[htb]
    \centering\small
    \caption{Page-language distribution of Khondo, per variant (all splits). Values are percentages
    of labeled pages. The benchmark is consistently Bangla-dominant with a substantial English minority.}
    \label{tab:lang-dist}
    \resizebox{0.7\columnwidth}{!}{
        \begin{tabular}{lcccc}
            \toprule
            Variant & Bangla & English & Mixed & Uncertain \\
            \midrule
            MonoSeq  & 68.6 & 23.7 & 5.8 & 1.9 \\
            MonoRand & 66.2 & 27.6 & 4.3 & 1.9 \\
            PolySeq  & 66.3 & 25.9 & 5.7 & 2.1 \\
            PolyInt  & 68.4 & 24.5 & 4.7 & 2.4 \\
            PolyRand & 64.0 & 27.9 & 5.6 & 2.5 \\
            \midrule
            All (avg.)      & 66.7 & 25.9 & 5.2 & 2.2 \\
            \bottomrule
        \end{tabular}
    }
\end{table}

\label{subsec:domain_dist}

\subsection{Ordering Bottleneck Experiment Details}

    \subsubsection{Prompt Sensitivity}
    \label{subsec:prompt_sense_appendix}
    
\begin{figure}[htb]
    \centering
    \includegraphics[width=\linewidth]{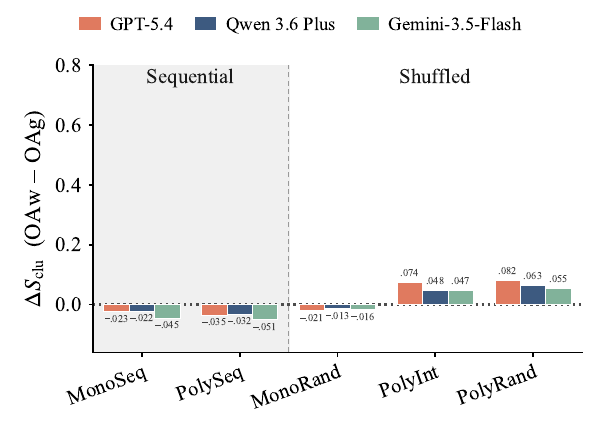}
    \caption{
        Change in clustering score,
        $\Delta S_{\mathrm{clu}} = S_{\mathrm{clu}}^{\mathrm{OAw}} - S_{\mathrm{clu}}^{\mathrm{OAg}}$, 
        for each model and packet variant, on the same scale as the ordering figure (\cref{fig:prompt-ord}).
        Where ordering swings by $0.43$--$0.67$ on the shuffled variants, clustering hardly moves: every bar stays within $0.09$ of zero. The order-aware prompt lowers clustering slightly on the sequential variants and MonoRand, and raises it slightly on PolyInt and PolyRand, so its effect is confined to page ordering rather than grouping. Full numbers in \Cref{tab:prompt}.
    }
    \label{fig:prompt-clu}
\end{figure}

    \Cref{fig:prompt-clu} shows the clustering counterpart of the ordering results on the same scale. Whereas ordering changes by $0.43$--$0.67$ on the shuffled variants, every clustering bar remains within $0.09$ of zero. This indicates that order-aware instructions have only a minimal effect on clustering compared with their substantial impact on ordering. The small residual changes are systematic rather than random noise. A paired permutation test confirms that both the upward and downward shifts across variants are statistically reliable and consistent across permutations.

    \Cref{tab:prompt} reports the complete results. In addition to the ordering and clustering effects shown in the preceding figures, it includes page-typing accuracy (PgAcc), which is largely unaffected by the prompt. PgAcc remains low across all settings (approximately $0.40$--$0.61$) and changes only marginally under OAw, with no consistent trend. This suggests that identifying a document's domain is difficult regardless of whether the model is instructed to preserve page order, reinforcing that correctly grouping pages does not imply correctly classifying their types. The only notable exception is Gemini-3.5-Flash, whose page-typing accuracy decreases by roughly $0.04$ on the poly variants under OAw, while the other two models remain essentially unchanged.

    \begin{table}[hbt]
    \centering\small
    \setlength{\tabcolsep}{4pt}
    \caption{
        Prompt sensitivity, full results. Order-agnostic (OAg) versus order-aware (OAw) instruction on identical packets ($N{\approx}100$ per cell, paired). The OAw prompt states that a document's pages may be presented out of order and must be re-sequenced; OAg prompt does not contain any order specific instruction. $S_{\mathrm{ord}}$ and $S_{\mathrm{clu}}$ are the ordering and clustering scores; PgAcc is page-typing accuracy (fraction of pages given the correct domain). $\Delta = \text{OAw}-\text{OAg}$; \textcolor{deltapos}{blue}/\textcolor{deltaneg}{orange} mark positive/negative $\Delta$.
    }
    \label{tab:prompt}
    \resizebox{\columnwidth}{!}{
        \begin{tabular}{llccc ccc ccc}
              \toprule
              & & \multicolumn{3}{c}{$S_{\text{ord}}$} & \multicolumn{3}{c}{$S_{\text{clu}}$} & \multicolumn{3}{c}{PgAcc} \\
              \cmidrule(lr){3-5}\cmidrule(lr){6-8}\cmidrule(lr){9-11}
              Variant & Model & OAg & OAw & $\Delta$ & OAg & OAw & $\Delta$ & OAg & OAw & $\Delta$ \\
              \midrule
              \multirow{3}{*}{MonoSeq}
              & GPT-5.4          & 0.909 & 0.839 & \dneg{$-$0.070} & 0.839 & 0.816 & \dneg{$-$0.023} & 0.564 & 0.545 & \dneg{$-$0.019} \\
              & Qwen 3.6 Plus    & 0.983 & 0.899 & \dneg{$-$0.084} & 0.886 & 0.864 & \dneg{$-$0.022} & 0.592 & 0.609 & \dpos{0.017} \\
              & Gemini-3.5-Flash & 0.998 & 0.920 & \dneg{$-$0.079} & 0.904 & 0.858 & \dneg{$-$0.045} & 0.529 & 0.540 & \dpos{0.011} \\
              \midrule
              \multirow{3}{*}{PolySeq}
              & GPT-5.4          & 0.856 & 0.759 & \dneg{$-$0.096} & 0.899 & 0.864 & \dneg{$-$0.035} & 0.519 & 0.527 & \dpos{0.008} \\
              & Qwen 3.6 Plus    & 0.996 & 0.938 & \dneg{$-$0.057} & 0.949 & 0.917 & \dneg{$-$0.032} & 0.561 & 0.501 & \dneg{$-$0.059} \\
              & Gemini-3.5-Flash & 0.988 & 0.937 & \dneg{$-$0.051} & 0.932 & 0.881 & \dneg{$-$0.051} & 0.546 & 0.503 & \dneg{$-$0.043} \\
              \midrule[1.5pt]
              \multirow{3}{*}{MonoRand}
              & GPT-5.4          & 0.049 & 0.602 & \dpos{$+$0.553} & 0.833 & 0.812 & \dneg{$-$0.021} & 0.559 & 0.574 & \dpos{0.015} \\
              & Qwen 3.6 Plus    & 0.082 & 0.592 & \dpos{$+$0.510} & 0.859 & 0.847 & \dneg{$-$0.013} & 0.590 & 0.600 & \dpos{0.010} \\
              & Gemini-3.5-Flash & 0.123 & 0.793 & \dpos{$+$0.670} & 0.857 & 0.842 & \dneg{$-$0.016} & 0.514 & 0.527 & \dpos{0.013} \\
              \midrule
              \multirow{3}{*}{PolyInt}
              & GPT-5.4          & 0.059 & 0.485 & \dpos{$+$0.426} & 0.641 & 0.715 & \dpos{0.074} & 0.404 & 0.425 & \dpos{0.021} \\
              & Qwen 3.6 Plus    & 0.128 & 0.614 & \dpos{$+$0.486} & 0.765 & 0.813 & \dpos{0.048} & 0.491 & 0.495 & \dpos{0.004} \\
              & Gemini-3.5-Flash & 0.079 & 0.698 & \dpos{$+$0.619} & 0.728 & 0.775 & \dpos{0.047} & 0.507 & 0.465 & \dneg{$-$0.042} \\
              \midrule
              \multirow{3}{*}{PolyRand}
              & GPT-5.4          & 0.035 & 0.542 & \dpos{$+$0.507} & 0.623 & 0.705 & \dpos{0.082} & 0.451 & 0.456 & \dpos{0.005} \\
              & Qwen 3.6 Plus    & 0.119 & 0.579 & \dpos{$+$0.460} & 0.765 & 0.829 & \dpos{0.063} & 0.513 & 0.536 & \dpos{0.024} \\
              & Gemini-3.5-Flash & 0.190 & 0.681 & \dpos{$+$0.490} & 0.710 & 0.765 & \dpos{0.055} & 0.540 & 0.504 & \dneg{$-$0.037} \\
              \bottomrule
          \end{tabular}
    }
\end{table}

    \subsubsection{Cross-Lingual Effects}
    \Cref{tab:xling} reports the per-cell results for all three tasks: ordering, clustering, and page typing. Ordering consistently favors English across every model and variant, although the magnitude of the gap varies. Clustering also favors English, but by a much smaller and more consistent margin. In contrast, page-typing accuracy exhibits a less consistent pattern. The gap is generally positive, with English pages classified more accurately, but it is noticeable primarily on the poly variants and weak or non-significant on the mono variants. Gemini-3.5-Flash is the only exception, achieving slightly higher page-typing accuracy on Bangla pages for the mono variants. Overall, page typing appears to be only weakly influenced by language, echoing its limited sensitivity to the ordering prompt observed in \Cref{tab:prompt}.

    \begin{table}[htb]
    \centering\small
    \setlength{\tabcolsep}{4pt}
    \caption{
        Cross-lingual contrast, full results. Size-matched monolingual \textbf{Bn} (Bangla) and \textbf{En} (English) packets, each assembled from single-language source forms with the packet-size distribution held fixed ($N\approx100$ per cell). $S_{\mathrm{ord}}$ and $S_{\mathrm{clu}}$ are the ordering and clustering scores; PgAcc is page-typing accuracy (fraction of pages given the correct domain). $\Delta=\text{En}-\text{Bn}$; \textcolor{deltapos}{blue}/\textcolor{deltaneg}{orange} mark positive/negative $\Delta$.
    }
    \label{tab:xling}

    \resizebox{\columnwidth}{!}{
        \begin{tabular}{clccc ccc ccc}
            \toprule
            & & \multicolumn{3}{c}{$s_{\mathrm{ord}}$} & \multicolumn{3}{c}{$s_{\mathrm{clu}}$} & \multicolumn{3}{c}{PgAcc} \\
            \cmidrule(lr){3-5}\cmidrule(lr){6-8}\cmidrule(lr){9-11}
            Variant & Model & Bn & En & $\Delta$ & Bn & En & $\Delta$ & Bn & En & $\Delta$ \\
            \midrule
            \multirow{3}{*}{MonoSeq}
            & GPT-5.4          & 0.704 & 0.981 & \dpos{0.277} & 0.887 & 0.952 & \dpos{0.065} & 0.360 & 0.464 & \dpos{0.104} \\
            & Qwen 3.6 Plus    & 0.931 & 1.000 & \dpos{0.069} & 0.895 & 0.962 & \dpos{0.068} & 0.435 & 0.538 & \dpos{0.103} \\
            & Gemini-3.5-Flash & 0.923 & 0.999 & \dpos{0.076} & 0.894 & 0.944 & \dpos{0.050} & 0.409 & 0.374 & \dneg{$-$0.035} \\
            \midrule
            \multirow{3}{*}{PolySeq}
            & GPT-5.4          & 0.744 & 0.979 & \dpos{0.235} & 0.905 & 0.962 & \dpos{0.056} & 0.338 & 0.470 & \dpos{0.133} \\
            & Qwen 3.6 Plus    & 0.902 & 1.000 & \dpos{0.098} & 0.909 & 0.965 & \dpos{0.056} & 0.306 & 0.524 & \dpos{0.219} \\
            & Gemini-3.5-Flash & 0.896 & 1.000 & \dpos{0.104} & 0.900 & 0.954 & \dpos{0.054} & 0.366 & 0.445 & \dpos{0.080} \\
            \midrule[1.5pt]
            \multirow{3}{*}{MonoRand}
            & GPT-5.4          & 0.660 & 0.905 & \dpos{0.245} & 0.885 & 0.946 & \dpos{0.061} & 0.380 & 0.443 & \dpos{0.063} \\
            & Qwen 3.6 Plus    & 0.555 & 0.913 & \dpos{0.357} & 0.883 & 0.938 & \dpos{0.055} & 0.437 & 0.513 & \dpos{0.075} \\
            & Gemini-3.5-Flash & 0.849 & 0.978 & \dpos{0.130} & 0.874 & 0.955 & \dpos{0.081} & 0.400 & 0.298 & \dneg{$-$0.102} \\
            \midrule
            \multirow{3}{*}{PolyInt}
            & GPT-5.4          & 0.648 & 0.920 & \dpos{0.272} & 0.855 & 0.925 & \dpos{0.070} & 0.356 & 0.454 & \dpos{0.098} \\
            & Qwen 3.6 Plus    & 0.598 & 0.890 & \dpos{0.291} & 0.878 & 0.926 & \dpos{0.048} & 0.337 & 0.513 & \dpos{0.176} \\
            & Gemini-3.5-Flash & 0.835 & 0.981 & \dpos{0.146} & 0.887 & 0.948 & \dpos{0.061} & 0.347 & 0.450 & \dpos{0.102} \\
            \midrule
            \multirow{3}{*}{PolyRand}
            & GPT-5.4          & 0.662 & 0.870 & \dpos{0.208} & 0.859 & 0.897 & \dpos{0.038} & 0.375 & 0.463 & \dpos{0.088} \\
            & Qwen 3.6 Plus    & 0.516 & 0.826 & \dpos{0.310} & 0.878 & 0.921 & \dpos{0.043} & 0.351 & 0.529 & \dpos{0.177} \\
            & Gemini-3.5-Flash & 0.826 & 0.975 & \dpos{0.148} & 0.890 & 0.939 & \dpos{0.019} & 0.374 & 0.393 & \dpos{0.019} \\
            \bottomrule
        \end{tabular}
    }
\end{table}

    \subsection{Form Language Classification}
    \label{subsec:lang_label}

\begin{figure*}[hbt]
    \centering

    \includegraphics[width=0.85\linewidth]{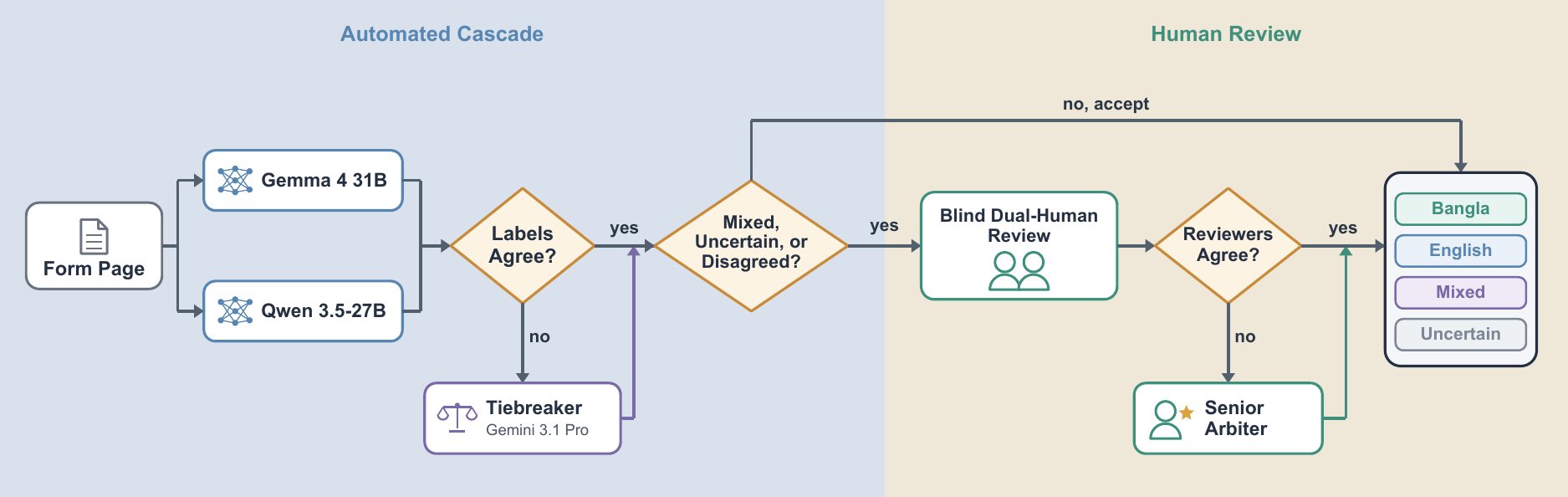}
    \caption{ Per-page language labeling/classifying pipeline. An automated cascade labels each form page: two small MLLMs (Gemma 4 31B, Qwen 3.5-27B) classify it independently,
    and pages on which they agree are accepted directly, while disagreements are resolved by a frontier tiebreaker (Gemini 3.1 Pro). Pages that are mixed, uncertain, or contested are routed to a blind dual-human review, with reviewer disagreements settled by an expert arbiter. Both paths converge on one of four labels: Bangla, English, Mixed, or Uncertain.}
    \label{fig:lang-pipeline}
\end{figure*}

    Each page is assigned one of four language labels: \emph{Bangla}, \emph{English}, \emph{Mixed}, or \emph{Uncertain}. Labels are produced through a multi-stage pipeline with human review of ambiguous cases (\Cref{fig:lang-pipeline}). Two open-weight MLLMs, \textbf{Gemma 4 31B} and \textbf{Qwen 3.5-27B}, classify each page independently. They achieve near-perfect agreement (Cohen's $\kappa = 0.97$), allowing $98.4\%$ of the $3{,}488$ pages to be labeled automatically. Where the models disagree, we resolve using a frontier MLLM, \textbf{Gemini 3.1 Pro Preview}. Pages labeled as \emph{Mixed} or \emph{Uncertain}, together with all pages on which the two smaller models disagree ($243$ pages, approximately $7\%$), are then reviewed by two independent human annotators. The $15$ pages on which the annotators disagree are resolved by a third expert arbiter.

    Human annotation provides the primary guarantee of label quality. On the subset of ambiguous pages requiring review, the two annotators achieve strong agreement (Cohen's $\kappa = 0.90$), with the few remaining disagreements resolved by a third expert. The models' higher agreement ($\kappa = 0.97$) is not directly comparable, as it is measured over the full dataset, where most pages are straightforward to classify. In contrast, the human agreement is measured only on the deliberately difficult cases, making it the stronger indicator of annotation reliability. The models' high agreement enables the vast majority of pages to be labeled automatically.

\begin{figure*}[t]
    \centering
    \resizebox{0.9\textwidth}{!}{
        \begin{minipage}[b]{0.31\linewidth}\centering
            \fbox{\includegraphics[width=\linewidth]{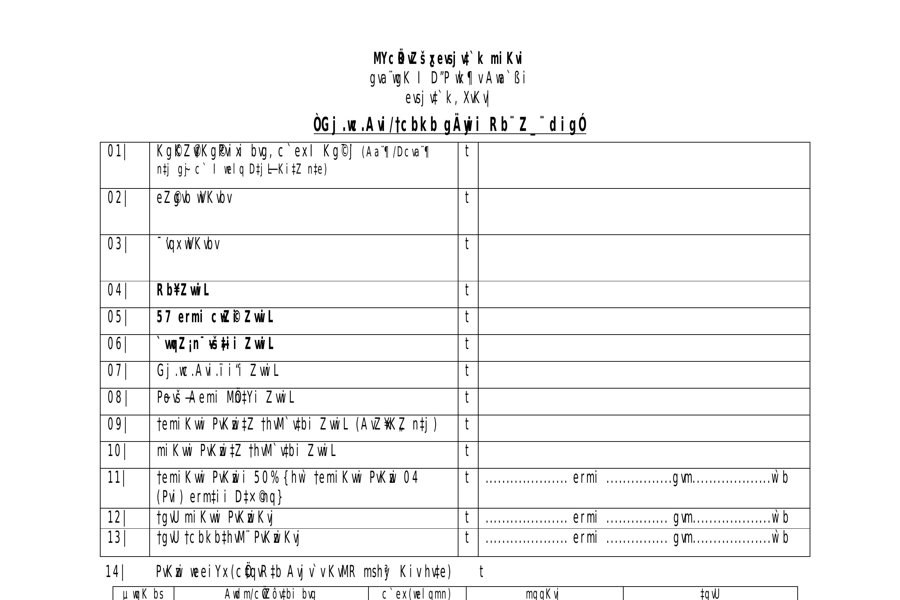}}\\[2pt]
            {\footnotesize (a) Legacy-encoded form}
        \end{minipage}\hfill
        \begin{minipage}[b]{0.31\linewidth}\centering
            \fbox{\includegraphics[width=\linewidth]{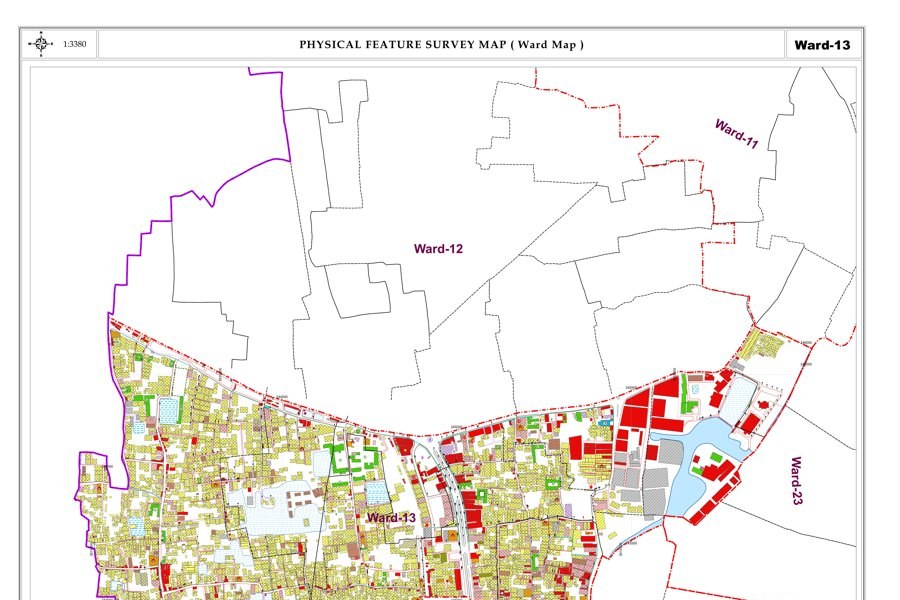}}\\[2pt]
            {\footnotesize (b) Figure-only page}
        \end{minipage}\hfill
        \begin{minipage}[b]{0.31\linewidth}\centering
            \fbox{\includegraphics[width=\linewidth]{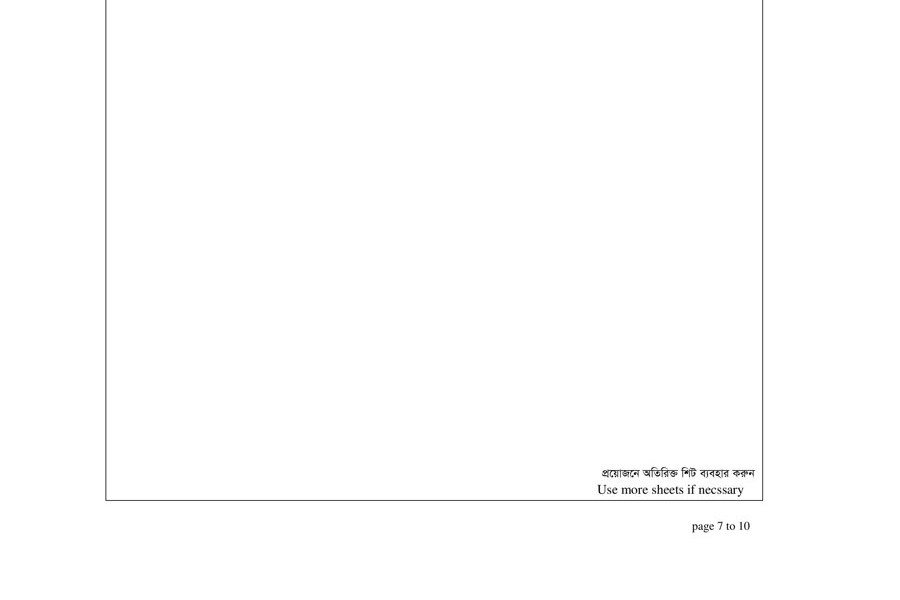}}\\[2pt]
            {\footnotesize (c) Near-empty page}
        \end{minipage}
    }
    \caption{
        Examples of pages labeled \textbf{Uncertain}. (a) A Bengali government form in a legacy, non-Unicode font (the Bijoy/SutonnyMJ family) rather than a Unicode-compatible method such as Avro), so it's rendered as unreadable text. (b) A ward map filed within a city-corporation specific application packet, that has no key-value pairs. (c) A near-empty continuation sheet with only a header and footer. Such pages hold little or no readable text yet are integral parts of their documents, so they are kept in the packets rather than discarded.
    }
    \label{fig:uncertain}
\end{figure*}

    Approximately $2\%$ of pages contain no field keys readable in either Bangla or English and are therefore labeled \emph{Uncertain}. These include (a) genuine government forms written in legacy non-Unicode Bangla fonts (e.g., Bijoy/SutonnyMJ), (b) pages dominated by maps or diagrams, and (c) blank or near-empty scans (\Cref{fig:uncertain}). Legacy forms remain in active use, and without the original fonts their text appears as unreadable glyphs despite being valid documents. We retain these pages because they are part of authentic document packets. Maps, diagrams, and separator pages provide meaningful context, and packet-splitting models must still assign them to the correct document and position. Removing such pages would fragment real packets and reduce the benchmark's realism. Since they are rare and verified through human review, they do not materially affect the clustering, ordering, or domain ground truth.

    \begin{figure*}[bht]
    
\begin{promptbox}{Language Classification Prompt}
You are a document language classifier. Given an image of a single form page, output the page's language using exactly one of these four labels:

LABELS:
- english: The form's field labels (keys) are written in English. Incidental content in Bangla --- logos, stamps, form codes, page headers, footers, scattered Bangla words inside English text --- does NOT change this label.
- bangla: The form's field labels (keys) are written in Bangla. Incidental content in English --- logos, stamps, form codes, page headers, footers, scattered English words inside Bangla text --- does NOT change this label.
- mixed: Most field labels are written in BOTH English and Bangla together (e.g., "Name / (*\bnimg{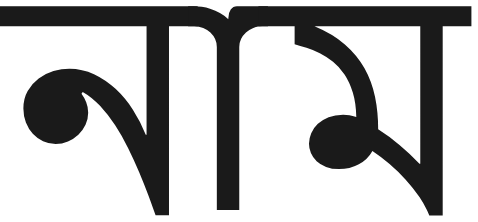}*)", "(*\bnimg{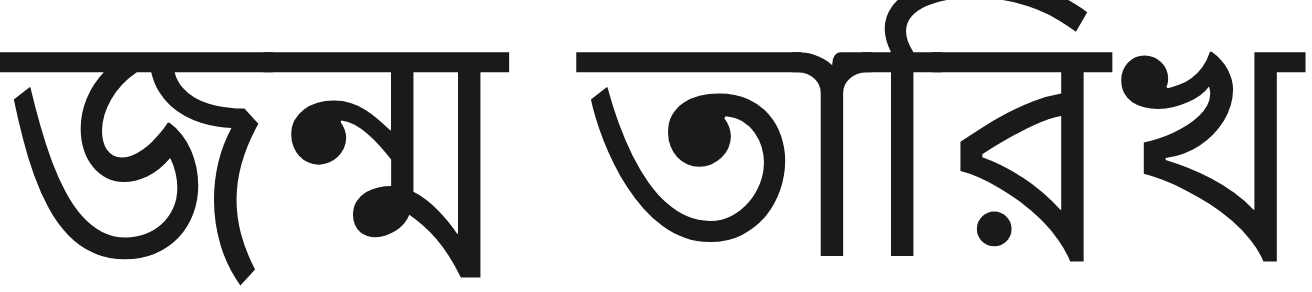}*) / Date of Birth"). The bilingual pairing must be the dominant pattern across visible field keys, not just a title or one or two isolated fields. The order of the two languages does not matter.
- not_sure: The page is blank, illegible, image- or figure-only, has no readable text, or you cannot confidently apply any of the rules above.

DECISION RULE:
- Look at the field labels (keys) used to introduce form fields, NOT the page title, header, footer, or instructional prose.
- A few stray words in the other language DO NOT trigger "mixed", in either direction.
- "mixed" requires bilingual key pairs (the same key written once in each language, in either order) to be the dominant pattern across the visible fields on
  the page.

OUTPUT FORMAT (exactly one JSON object, no surrounding text, no markdown fences):
{
    "language": "<english / bangla / mixed / not_sure>",
    "confidence": <float between 0.0 and 1.0>,
    "evidence": "<one short snippet of text copied verbatim from the page (max ~80 chars), or a brief reason such as 'blank scan' or 'image only' if not_sure>"
}

RULES:
- Output exactly ONE JSON object. No arrays, no extra text, no markdown fences, no commentary.
- "evidence" must be a real text snippet from the page (or a short description for not_sure).
- "confidence" reflects how sure you are after applying the decision rule above.

Image: <image>
Your verdict:
\end{promptbox}
\caption{Prompt used to classify forms into English/Bangla/Mixed/Unsure language groups. See \Cref{fig:lang-pipeline} for more details.}
\label{prompt:lang-clf}
\end{figure*}

\subsection{Effects of Model Size Scaling on Finetuning}

To examine whether supervised fine-tuning (SFT) can narrow the performance gap between open-weight and flagship MLLMs for packet splitting, we fine-tuned Qwen3-VL at model sizes ranging from 4B to 32B. We selected Qwen3-VL because it achieved the strongest overall zero-shot performance in our SFT experiments (\Cref{tab:sft}). Our results show that \emph{scaling provides the greatest benefit on the most challenging variants}. On the three shuffled variants, performance improves monotonically with model size, consistently following the trend $32\text{B} > 8\text{B} > 4\text{B}$ (e.g., PolyRand: $0.559 \rightarrow 0.636 \rightarrow 0.712$). In contrast, the two sequential variants are already close to saturation at 8B, leaving little room for further gains at 32B (e.g., MonoSeq: $0.853 \rightarrow 0.849$). Detailed results are provided in \Cref{tab:scale}.

\begin{figure*}[htb] 
    \centering
    \includegraphics[width=0.7\linewidth]{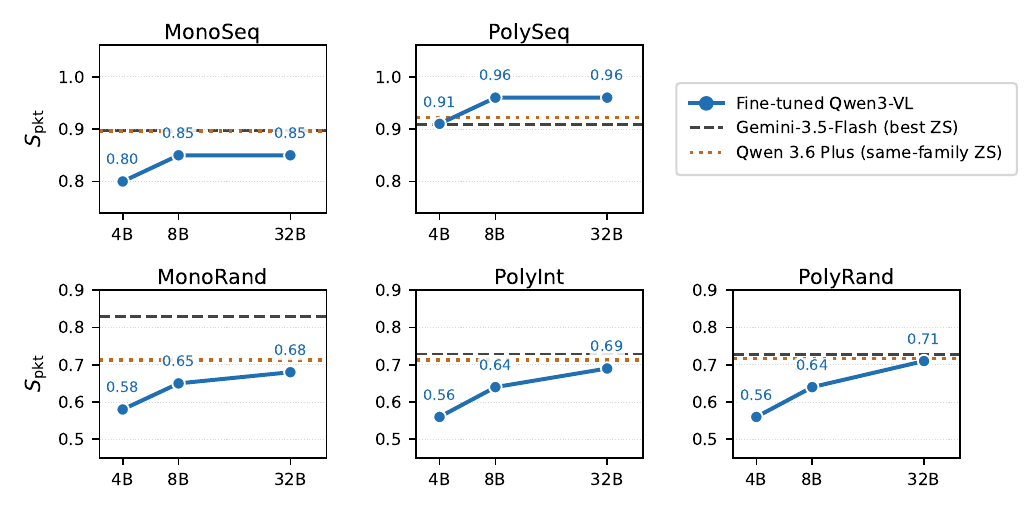}
    \caption{
        Packet score $S_{\mathrm{pkt}}$ of fine-tuned Qwen3-VL (using order-aware prompt, OAw) against backbone sizes 4B, 8B, and 32B. One panel per concatenation variant (top row: sequential; bottom row: shuffled). The solid blue curve is for fine-tuned Qwen3-VL; the dashed and dotted lines are for training-free inferences of Gemini-3.5-Flash (best zero-shot) and Qwen 3.6 Plus (same-family zero-shot) respectively, which coincide on MonoSeq. Relative to these references, the fine-tuned model approaches them on MonoSeq and exceeds both on PolySeq, while on the shuffled variants it climbs steadily but remains below them.
    }
    \label{fig:scaling}
\end{figure*}

The reduction in the performance gap is not uniform across variants. On the sequential variants, the fine-tuned models approach flagship MLLM performance, with the 32B model outperforming both \textbf{Gemini-3.5-Flash} and \textbf{Qwen 3.6 Plus} on PolySeq. On the shuffled variants, scaling substantially narrows the gap but does not eliminate it at 32B, suggesting that additional model capacity or training may yield further improvements. Ordering remains the slowest to improve, consistent with the ordering bottleneck observed in both the training-free and prompt-sensitivity experiments. By contrast, page domain classification shows no consistent improvement with increasing model size and remains relatively low across all scales (\Cref{tab:scale}), mirroring the persistent difficulty of page typing observed throughout the benchmark. The per-variant scaling trends are illustrated in \Cref{fig:scaling}.

\begin{table*}[htb]
    \centering\small
    \setlength{\tabcolsep}{6pt}
    \caption{Effect of model size on fine-tuned Qwen3-VL, using the order-aware (OAw) prompt on the Khondo test split (210 packets per variant). \emph{Base} is the pre-fine-tuning model and \emph{FT} the fine-tuned model. $S_{\mathrm{clu}}$, $S_{\mathrm{ord}}$, $S_{\mathrm{pkt}}$, and PgAcc are defined as in \Cref{tab:zeroshot}. Fine-tuning improves most metrics over the base model, and clustering rises with model size on every variant while ordering and packet scores rise on the shuffled variants, whereas page typing (PgAcc) stays low regardless of size.}
    \label{tab:scale}
    
    \resizebox{0.8\textwidth}{!}{
        \begin{tabular}{ll ccc ccc ccc ccc}
            \toprule
            & & \multicolumn{3}{c}{$S_{\mathrm{clu}}$} & \multicolumn{3}{c}{$S_{\mathrm{ord}}$}
            & \multicolumn{3}{c}{$S_{\mathrm{pkt}}$} & \multicolumn{3}{c}{PgAcc} \\
            \cmidrule(lr){3-5}\cmidrule(lr){6-8}\cmidrule(lr){9-11}\cmidrule(lr){12-14}
            \textbf{Variant} & & 4B & 8B & 32B & 4B & 8B & 32B & 4B & 8B & 32B & 4B & 8B & 32B \\
            \midrule
            \multirow{2}{*}{MonoSeq}  & Base & 0.591 & 0.662 & 0.845 & 0.612 & 0.655 & 0.897 & 0.601 & 0.658 & 0.871 & 0.297 & 0.273 & 0.360 \\
                                    & FT   & 0.843 & 0.873 & 0.878 & 0.761 & 0.832 & 0.819 & 0.802 & 0.853 & 0.849 & 0.566 & 0.517 & 0.514 \\
            \cmidrule(l){1-14}
            \multirow{2}{*}{PolySeq}  & Base & 0.582 & 0.782 & 0.915 & 0.639 & 0.773 & 0.952 & 0.610 & 0.778 & 0.933 & 0.169 & 0.180 & 0.274 \\
                                    & FT   & 0.945 & 0.971 & 0.978 & 0.866 & 0.949 & 0.948 & 0.906 & 0.960 & 0.963 & 0.283 & 0.302 & 0.305 \\
            \midrule[1.2pt]
            \multirow{2}{*}{MonoRand} & Base & 0.515 & 0.596 & 0.769 & 0.023 & 0.089 & 0.187 & 0.269 & 0.342 & 0.478 & 0.271 & 0.268 & 0.349 \\
                                    & FT   & 0.840 & 0.875 & 0.889 & 0.316 & 0.430 & 0.465 & 0.578 & 0.652 & 0.677 & 0.528 & 0.510 & 0.635 \\
            \cmidrule(l){1-14}
            \multirow{2}{*}{PolyInt}  & Base & 0.300 & 0.403 & 0.652 & $-$0.009 & 0.067 & 0.150 & 0.146 & 0.235 & 0.401 & 0.146 & 0.125 & 0.213 \\
                                    & FT   & 0.788 & 0.819 & 0.872 & 0.333 & 0.455 & 0.509 & 0.561 & 0.637 & 0.691 & 0.270 & 0.295 & 0.307 \\
            \cmidrule(l){1-14}
            \multirow{2}{*}{PolyRand} & Base & 0.261 & 0.419 & 0.658 & 0.033 & 0.097 & 0.149 & 0.147 & 0.258 & 0.404 & 0.127 & 0.143 & 0.250 \\
                                    & FT   & 0.797 & 0.827 & 0.883 & 0.321 & 0.444 & 0.541 & 0.559 & 0.636 & 0.712 & 0.276 & 0.293 & 0.332 \\
            \bottomrule
        \end{tabular}
    }
\end{table*}

\onecolumn
\clearpage
\subsection{Case Studies}
\label{subsec:case_studies}

\begin{figure*}[htb]
    \centering
    \includegraphics[width=\linewidth]{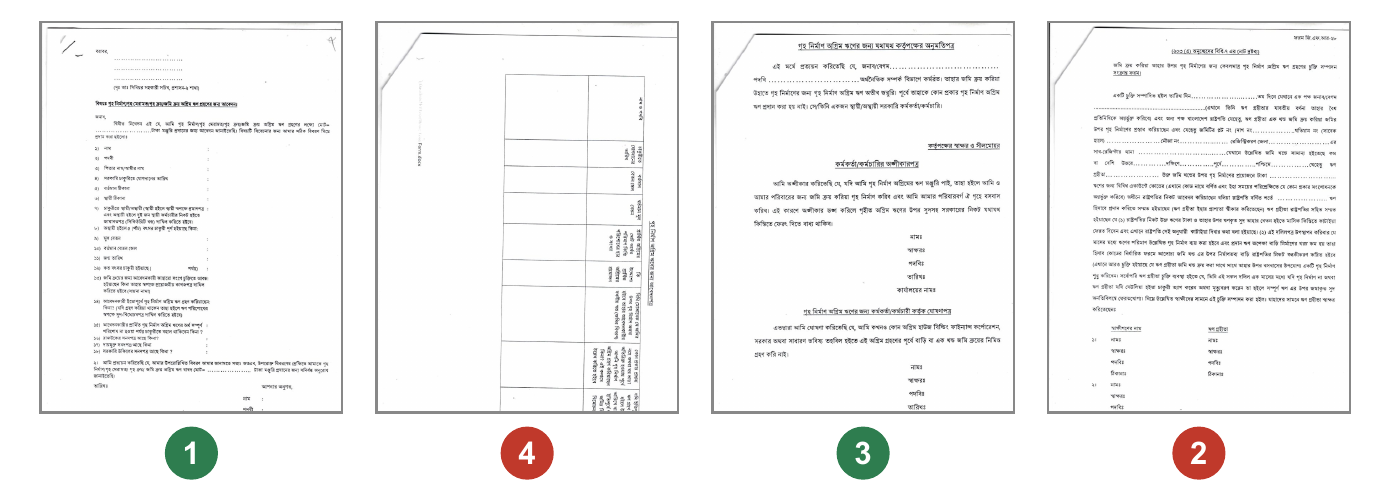}
    \caption{
        \textbf{Ordering bottleneck (Gemini 3.5 Flash):} Four pages of one Bangla employment form from a shuffled packet, shown in the model's output order. The badge under each page is its true position (green = correct slot, red = out of place): a correct prediction reads $1 \rightarrow 2 \rightarrow 3 \rightarrow 4$, but the model produces $1 \rightarrow 4 \rightarrow 3 \rightarrow 2$, swapping 2nd and 4th page of ground truth order. The model assigns the packet's pages to the correct forms but cannot reconstruct page order.
    }
    \label{fig:case-ordering}
\end{figure*}

\textbf{Correct grouping with incorrect page order:} In \Cref{fig:case-ordering}, Gemini-3.5-Flash correctly groups all four pages of an \textit{employment} form but returns them in the order $1 \rightarrow 4 \rightarrow 3 \rightarrow 2$. Although the pages are assigned to the correct form, the second and fourth pages are swapped, while the first and third remain in place. This single ordering error is sufficient to reduce $S_{\mathrm{ord}}$, even though $S_{\mathrm{clu}}$ remains high (Table~\ref{tab:zeroshot}). The model successfully identifies which pages belong together but fails to recover their correct reading order.

\begin{figure*}[t]
    \centering
    \includegraphics[width=\linewidth]{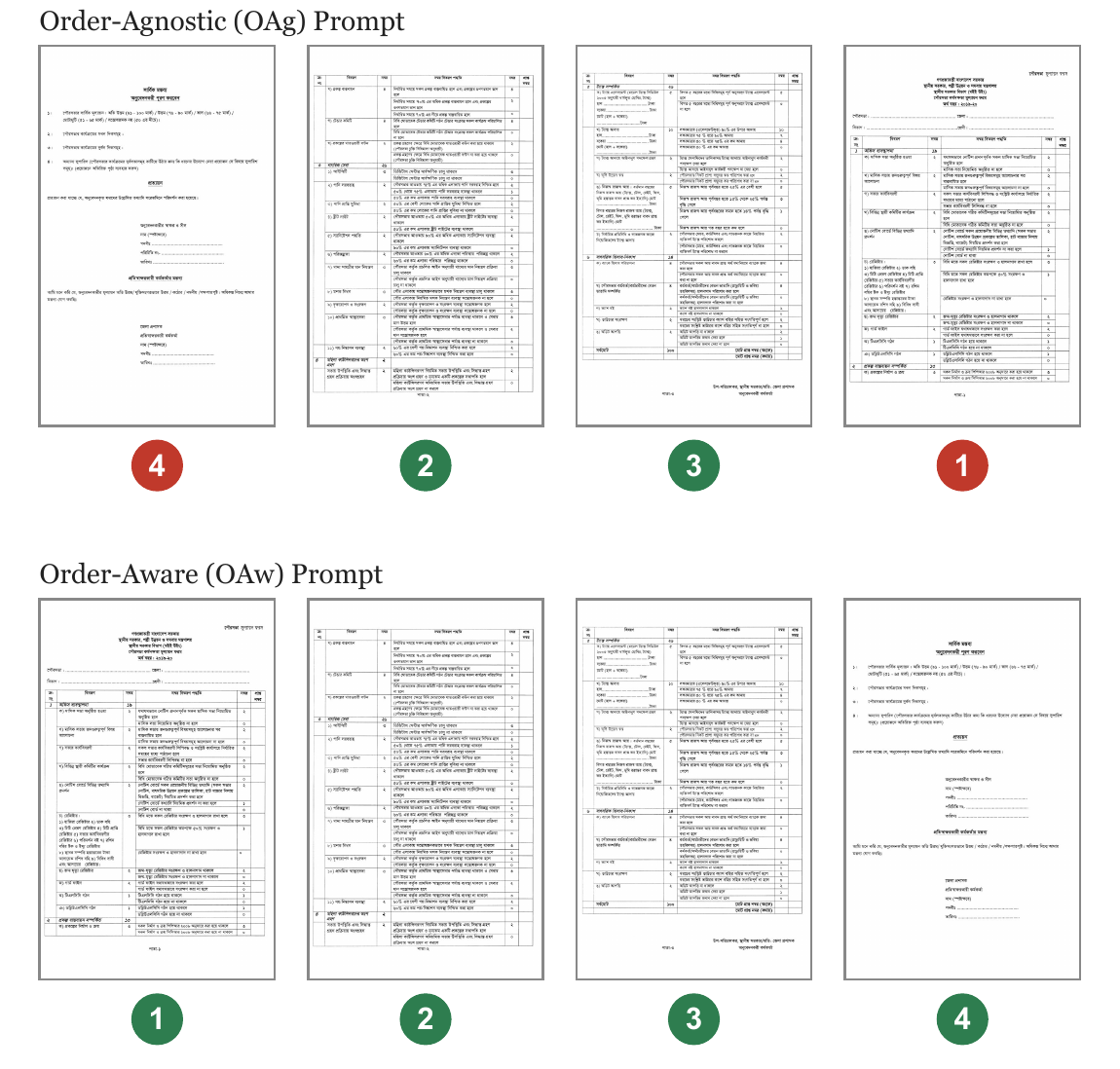}
    \caption{
        \textbf{Prompt sensitivity (Gemini 3.5 Flash):} Four pages of one Bangla local-government form from a shuffled packet, shown in the model's output order under two prompts. The badge under each page is its true position (green = correct slot, red = out of place). Under the order-agnostic (OAg) prompt the model reproduces the presented page order, reading $4 \rightarrow 2 \rightarrow 3 \rightarrow 1$; under the order-aware (OAw) prompt the same model re-sequences the pages to the correct order $1 \rightarrow 2 \rightarrow 3 \rightarrow 4$. The pages are identical across the two runs, so the order specific instruction in prompt alone drives the change.
    }
    \label{fig:case-prompt}
\end{figure*}

\textbf{Impact of order-aware prompt instructions:} \Cref{fig:case-prompt} shows the same shuffled form evaluated with both an order-agnostic (OAg) and an order-aware (OAw) prompt. Under the order-agnostic prompt, the model preserves the presented sequence, producing $4 \rightarrow 2 \rightarrow 3 \rightarrow 1$. With the order-aware prompt, it correctly reconstructs the original order, $1 \rightarrow 2 \rightarrow 3 \rightarrow 4$. Because the pages are identical across both runs, the difference is attributable solely to the prompt. Without an explicit instruction to reorder pages, the model simply follows the input sequence rather than inferring the correct reading order. This behavior is systematic: across the benchmark, the order-aware prompt recovers much of the ordering accuracy lost under the order-agnostic prompt, although not all of it (\Cref{fig:prompt-ord}).

\begin{figure*}[t]
    \centering
    \includegraphics[width=\linewidth]{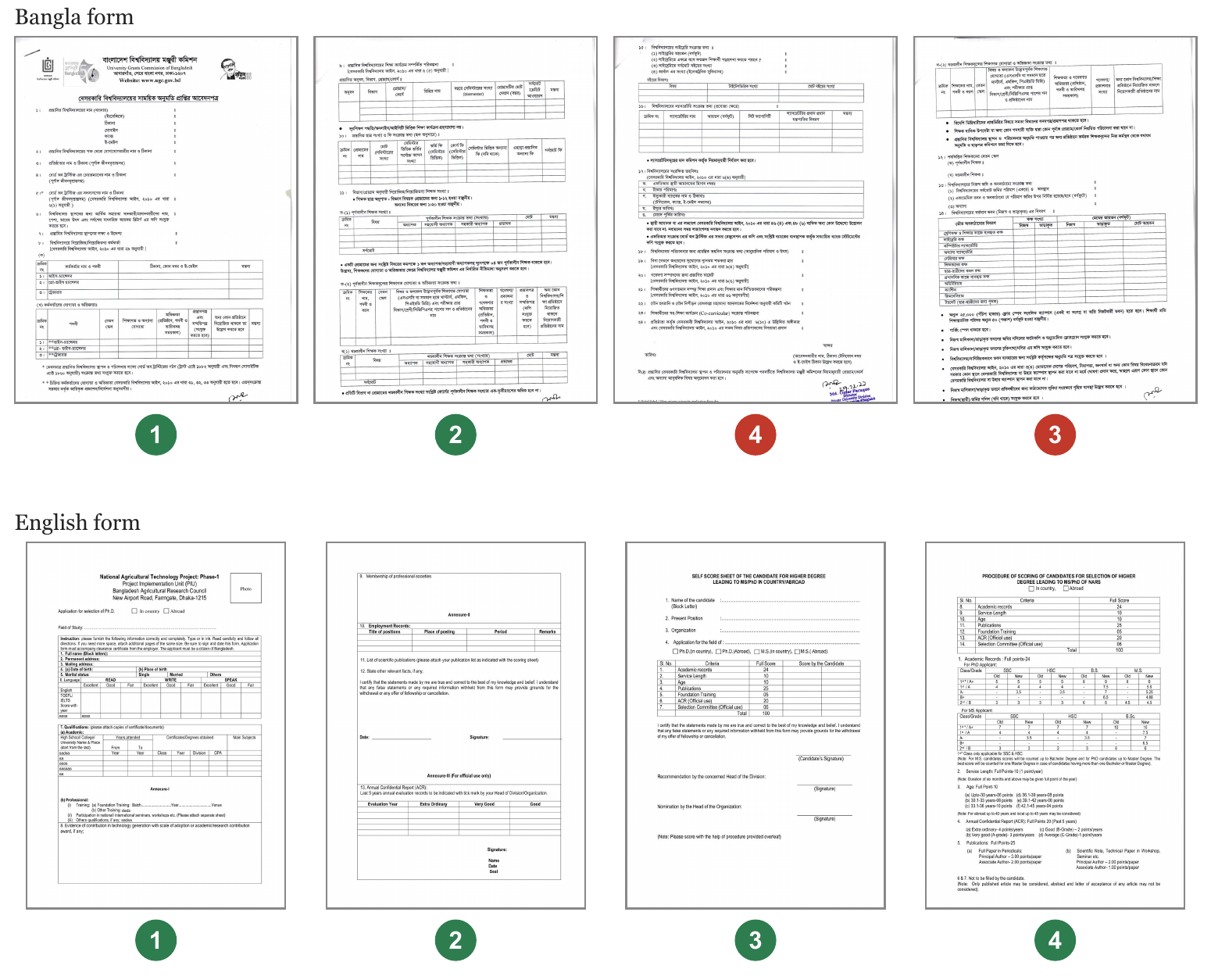}
    \caption{
        \textbf{Cross-lingual ordering gap (Qwen 3.6 Plus):} Two education application forms of the same length from shuffled packets, one Bangla (top) and one English (bottom), each shown in the model's output order. The badge under each page is its true position (green = correct slot, red = out of place). The model orders the English form correctly, $1 \rightarrow 2 \rightarrow 3 \rightarrow 4$, but misorders the Bangla one, $1 \rightarrow 2 \rightarrow 4 \rightarrow 3$. This English-over-Bangla ordering gap is systematic across models and variants in the cross-lingual experiment.
    }
    \label{fig:case-lang}
\end{figure*}

\textbf{Ordering is more reliable in English than in Bangla:} \Cref{fig:case-lang} compares two education application forms of equal length, one in English and one in Bangla. Qwen 3.6 Plus correctly groups both forms and orders the English form perfectly ($1 \rightarrow 2 \rightarrow 3 \rightarrow 4$), but misorders the Bangla form as $1 \rightarrow 2 \rightarrow 4 \rightarrow 3$, swapping the final two pages. Since the forms share the same domain, length, and shuffle pattern, the difference is attributable to language. This pattern is consistent across all evaluated models: ordering is more accurate on English packets than on Bangla packets in every case, with the largest gap observed for Qwen 3.6 Plus, which is the least reliable of the three strongest models at ordering Bangla forms (\Cref{tab:xling}).


\end{document}